\documentclass[runningheads]{llncs}

\usepackage{eccv}
\usepackage{eccvabbrv}
\usepackage{graphicx}
\usepackage{booktabs}
\usepackage{algorithm}
\usepackage{algorithmic}
\usepackage{bm}
\renewcommand{\vec}[1]{\bm{#1}}
\newcommand{\mat}[1]{\bm{#1}}
\newcommand*{\boldone}{\text{\usefont{U}{bbold}{m}{n}1}}
\usepackage{gensymb}
\usepackage[hang,flushmargin]{footmisc}
\usepackage{cuted}
\usepackage[accsupp]{axessibility}

\usepackage{hyperref}
\usepackage{orcidlink}

\begin{document}

\title{Beyond the Contact: Discovering Comprehensive Affordance for 3D Objects from Pre-trained 2D Diffusion Models} 

\titlerunning{Discovering Comprehensive Affordance for 3D Objects}

\author{Hyeonwoo Kim\inst{1*} \and
Sookwan Han\inst{1*} \and
Patrick Kwon\inst{2} \and
Hanbyul Joo\inst{1}
}

\authorrunning{H. Kim et al.}

\institute{Seoul National University \\
\and
Naver Webtoon AI\\
}

\maketitle
\def\thefootnote{*}\footnotetext{Indicates equal contribution}\def\thefootnote{\arabic{footnote}}

\begin{figure}
\centering
\vspace{-15pt}
\includegraphics[width=\linewidth, trim={0 0 0 0},clip]{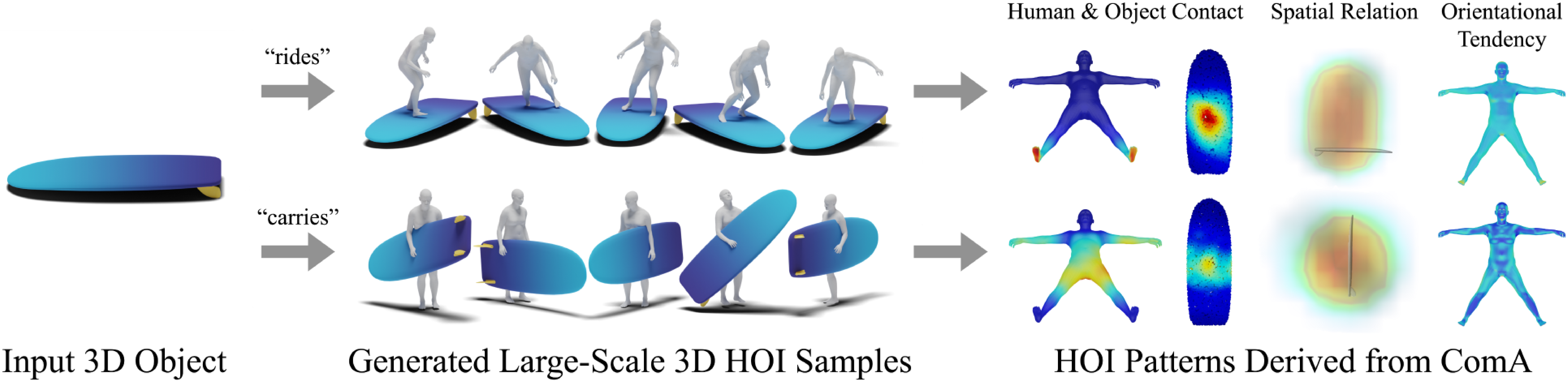}
\captionof{figure}{Given a 3D object, we generate numerous 3D Human-Object Interaction (HOI) samples using text prompts, and learn a novel affordance representation called \textit{Comprehensive Affordance} (ComA) which models both contact and non-contact HOI patterns.}
\label{fig:teaser}
\vspace{-30pt}
\end{figure}

\begin{abstract}
Understanding the inherent human knowledge in interacting with a given environment (\emph{e.g.}, affordance) is essential for improving AI to better assist humans.
While existing approaches primarily focus on human-object contacts during interactions, such affordance representation cannot fully address other important aspects of human-object interactions (HOIs), \emph{i.e.}, patterns of relative positions and orientations.
In this paper, we introduce a novel affordance representation, named \textit{Comprehensive Affordance} (ComA). Given a 3D object mesh, ComA models the distribution of relative orientation and proximity of vertices in interacting human meshes, capturing plausible patterns of contact, relative orientations, and spatial relationships.
To construct the distribution, we present a novel pipeline that synthesizes diverse and realistic 3D HOI samples given any 3D object mesh. The pipeline leverages a pre-trained 2D inpainting diffusion model to generate HOI images from object renderings and lifts them into 3D. To avoid the generation of false affordances, we propose a new inpainting framework, \textit{Adaptive Mask Inpainting}. Since ComA is built on synthetic samples, it can extend to any object in an unbounded manner.
Through extensive experiments, we demonstrate that ComA outperforms competitors that rely on human annotations in modeling contact-based affordance. Importantly, we also showcase the potential of ComA to reconstruct human-object interactions in 3D through an optimization framework, highlighting its advantage in incorporating both contact and non-contact properties.
\end{abstract}
\section{Introduction}
\label{sec:intro}

Humans possess an innate ability to perceive the functionalities provided by the environment or objects and efficiently utilize them---a concept known as affordance~\cite{jjgibson}. 
Affordance incorporates a variety of patterns typically observable in human-object interactions (HOIs), including not only plausible physical contact, but also spatial and orientational non-contact relationships. For example, while using a laptop, the orientation and distance patterns between human face and the screen can be considered part of affordance, alongside the hand regions touching the laptop keyboard. 
Teaching such affordance knowledge to machines is crucial for enabling them to better understand or mimic human behaviors. Consequently, numerous studies emerged to implement such affordance knowledge into AI and robotics~\cite{Robo-Affordance, LOCATE,IAGNet,POSA, DECO,objectpopup, Affordance-Insertion, Affordance-Diffusion, CHORUS}.

Most previous approaches however, primarily focus on a limited spectrum of affordances. 
Some approaches represent the human and object regions in contact using contact scores on 2D images~\cite{Robo-Affordance, LOCATE}, 3D objects~\cite{IAGNet}, and 3D humans~\cite{POSA, DECO} using direct supervision from human annotations.
Others represent affordances by inferring~\cite{objectpopup} or generating~\cite{Affordance-Insertion, Affordance-Diffusion} plausible human-object pairs in contact-oriented object categories. 
While a few recent studies tackle learning the spatial relations~\cite{CHORUS, LEMON} between a human and an object, these methods mainly focus on object categories where physical contact is dominant than non-contact patterns.

\begin{figure}[t]
\centering
\includegraphics[width=\columnwidth]{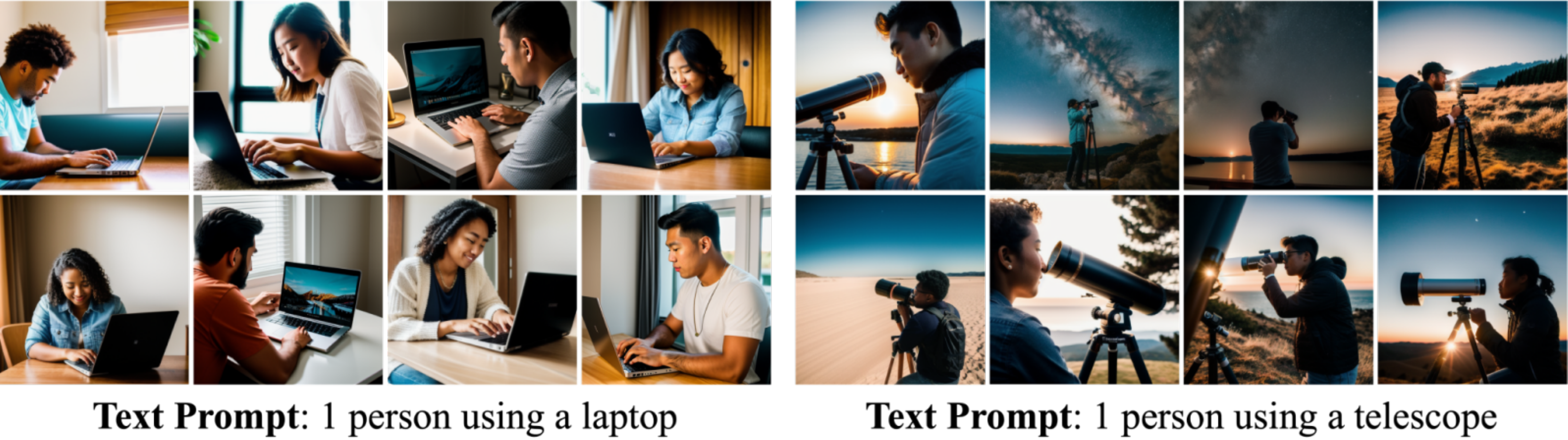}
\vspace{-15pt}
\caption{Typically, people (1) view the screen (2) from relatively distant distance while using a laptop (Left), whereas they (1) peer into it (2) from a close distance while using a telescope (Right). Pre-trained diffusion model has a knowledge of these (1) orientational and (2) spatial relation between human and object during interaction.} 
\label{fig:implicit_effects}
\vspace{-15pt}
\end{figure}

In this paper, we introduce \textit{Comprehensive Affordance} (ComA), a novel representation of an affordance encompassing the various aspects of human-object interaction, including orientational tendencies and relative positions.
The motivation for our design arises from the observation that there exist typical 3D relations between entire body parts and objects during HOIs.
For instance, human faces, torsos, and arms often exhibit specific distances and orientations with some variability, when interacting with certain object categories, as shown in  Fig.~\ref{fig:implicit_effects}. 
ComA is designed to model these patterns by constructing a distribution between each pair of object surface points and human surface points in 3D. This distribution represents the likelihood of human body parts displaying specific spatial and orientational relations with object points, as shown in Fig.~\ref{fig:overview}.

Learning ComA requires various 3D HOI samples showing the way humans use the target object in 3D, from which pairwise distributions of 3D HOI relations can be constructed. Manually annotating such cues is challenging, compared to the annotations focusing solely on contact regions~\cite{LOCATE, InterDiff, objectpopup, DECO, IAGNet, LEMON}.
As a key solution, we present a pipeline to synthesize a large-scale synthetic 3D HOI samples leveraging a pre-trained 2D diffusion model.
Interestingly, we observe that a pre-trained 2D diffusion model captures the affordance knowledge in the form of 2D images, as shown in Fig.~\ref{fig:implicit_effects}, where the images are synthesized from text prompts. However, leveraging these 2D HOI cues to construct ComA for the given 3D object (which requires 3D HOI samples) presents the following challenges: (1) the diffusion model needs to be applied for inserting plausible humans without altering the original shape and pose of the target object, and (2) 3D HOI samples should be lifted from the synthesized 2D images. To address these challenges, we present \textit{Adaptive Mask Inpainting} to insert humans without altering the appearance of target objects in 2D, along with a 3D lifting pipeline from 2D HOI cues.

We demonstrate the efficacy of our approach by learning ComA on 100 3D object meshes in diverse categories collected from various sources~\cite{BEHAVE, InterCap, ShapeNet, SAPIEN, PartNet, SketchFab}. 
We compare ComA with existing approaches~\cite{DECO, IAGNet} on the BEHAVE dataset~\cite{BEHAVE}, demonstrating that ours learned from synthetic cues outperforms competitors trained on manual annotations for contact-based affordance. 
Additionally, we verify the advantage of ComA against contact-only representations, in the scenario of reconstructing HOIs via an optimization framework.
We will release the results and our source code\footnote[1]{\url{https://github.com/snuvclab/coma}}.

In summary, our main contributions can be summarized as follows: (1) we introduce a new representation ComA, which models the distribution of both explicit contact and non-contact patterns during HOIs; (2) we present a pipeline to build ComA from numerous synthetic samples by leveraging a pre-trained 2D diffusion model, scalable to any 3D object without requiring laborious annotations; and (3) we propose Adaptive Mask Inpainting, a method for preserving the original context of an image when using inpainting diffusion model.
\section{Related Work}
\label{sec:related}

\noindent \textbf{Learning Visual Affordance.}
Affordance was introduced by Gibson~\cite{jjgibson} as a set of functionalities that the environment furnishes to an agent (\emph{e.g.}, human, animal). The concept extends to the vision and robotics area, where the focus is on teaching embodied agents to interact with scenes~\cite{Robo-Affordance, nagarajan2020learning,kPAM-SC,Levine_2022_Robotic_navigation, affordance_robotics_labeling}, mimicking human-object~\cite{hoi_3dreconhuman,BEHAVE,hoi_capture_infer_dense_body,hoi_contact_from_mono_video,hoi_couch,hoi_holisticscenepp,hoi_imapper,hoi_longterm_humanmotion,hoi_motion_from_mono_video,hoi_neuralfreeviewpointhoi,hoi_neuralhofusion,hoi_pigraphs,hoi_place,objectpopup,POSA}, or hand-object~\cite{hand_contactdb,hand_contactgrasp,hand_contactpose,hand_grab,hand_toch} interactions. Earlier methods focused on learning action category labels~\cite{cai2016understanding, Hermans2011AffordancePV, Lee2015PredictingIO} with bounding boxes but lacked a detailed description of affordance. Due to the limited affordance information available from text descriptions or labels, some studies depicted affordance as contact regions or heatmaps on objects~\cite{Robo-Affordance, LOCATE, IAGNet} and humans~\cite{DECO, POSA}.
However, learning affordances beyond the contact was challenging since existing datasets created through a multi-camera system~\cite{BEHAVE, InterCap}, motion capture~\cite{prox}, and manual process~\cite{jian2023affordpose, DECO,IAGNet,3daffordancenet} focused only on annotating contact information rather than capturing the holistic aspect of HOI. Following studies generate plausible HOI samples in 2D~\cite{Affordance-Diffusion, Affordance-Insertion} or 3D~\cite{PSI, InterDiff} which has the potential for learning such broader HOI knowledge, but do not address the problem directly. In contrast, our method generates 3D HOI samples and extracts ComA for modelling beyond the contact.

\noindent \textbf{Data Synthesis for Learning.} There have been approaches to supplement data in fields lacking efficient annotation process by leveraging generative models, irrespective of the research area. Many methods employ GANs \cite{gan, stylegan} or diffusion models \cite{Rombach_2022_CVPR, ddpm, ddim} to augment data \cite{data_generate_diffusion_1,data_generate_diffusion_2} for various vision tasks, including perception and representation learning \cite{data_generate_diffusion_3}, classification \cite{classification_data_generate_1,classification_data_generate_2,classification_data_generate_3},
segmentation \cite{segmentation_data_generate, segmentation_data_generate2, segmentation_data_generate3}, dense visual alignment \cite{gangealing}, and further extending to 3D tasks, such as neural rendering \cite{neural_rendering_dataset_generated, neural_rendering_dataset_generated_2}, shape reconstruction \cite{shape_recon_dataset_generated}, and so forth. The major challenge in leveraging synthesized data (or generative models for cues) is controllability, as the generator, even when conditioned, typically produces free samples. Ali \etal\cite{generate_viewpoint} presents a method to control the viewpoint of the synthesized image by modeling the viewpoint-free latent space. CHORUS \cite{CHORUS} applies cascaded filtering to control the quality of the generated dataset to learn 3D human-object spatial relations, similar to our approach.

\noindent \textbf{Diffusion Model for Synthesizing HOI Images.}
While diffusion models~\cite{sohl2015deep_nonequilibriumthermodynamics, ddpm, ddim} excel at generating realistic images~\cite{Rombach_2022_CVPR, dalle2}, few address the generation of 2D human-object pairs (\emph{i.e.}, HOI images). Inpainting models~\cite{Rombach_2022_CVPR, RePaint}, while a common choice for human insertion, often compromise scene details, resulting in images with false affordances. Image editing techniques~\cite{SDEdit, Prompt2Prompt, null_text_inversion} may offer an alternative but tend to prioritize style changes and struggle with creating new geometry. 
Recently, Ye \etal~\cite{Affordance-Diffusion} employs an additional layout network to determine the inpainting region, and Kulal \etal~\cite{Affordance-Insertion} trains a diffusion model on video clips to generate HOI images. 
In contrast to these approaches, we create HOI images by inpainting humans while maintaining the context of the original image, without additional network training.
Concurrent work by Li \etal~\cite{genzi} introduces \textit{Dynamic Mask Inpainting}, which uses attention masks to change inpainting masks over timesteps to insert human into a rendered 3D scene, while we explicitly apply segmentation model to adapt the inpainting mask.
\section{Method}
\label{sec:method}

\begin{figure}[t]\centering
\includegraphics[width=\linewidth, trim={0 0 0 0},clip]{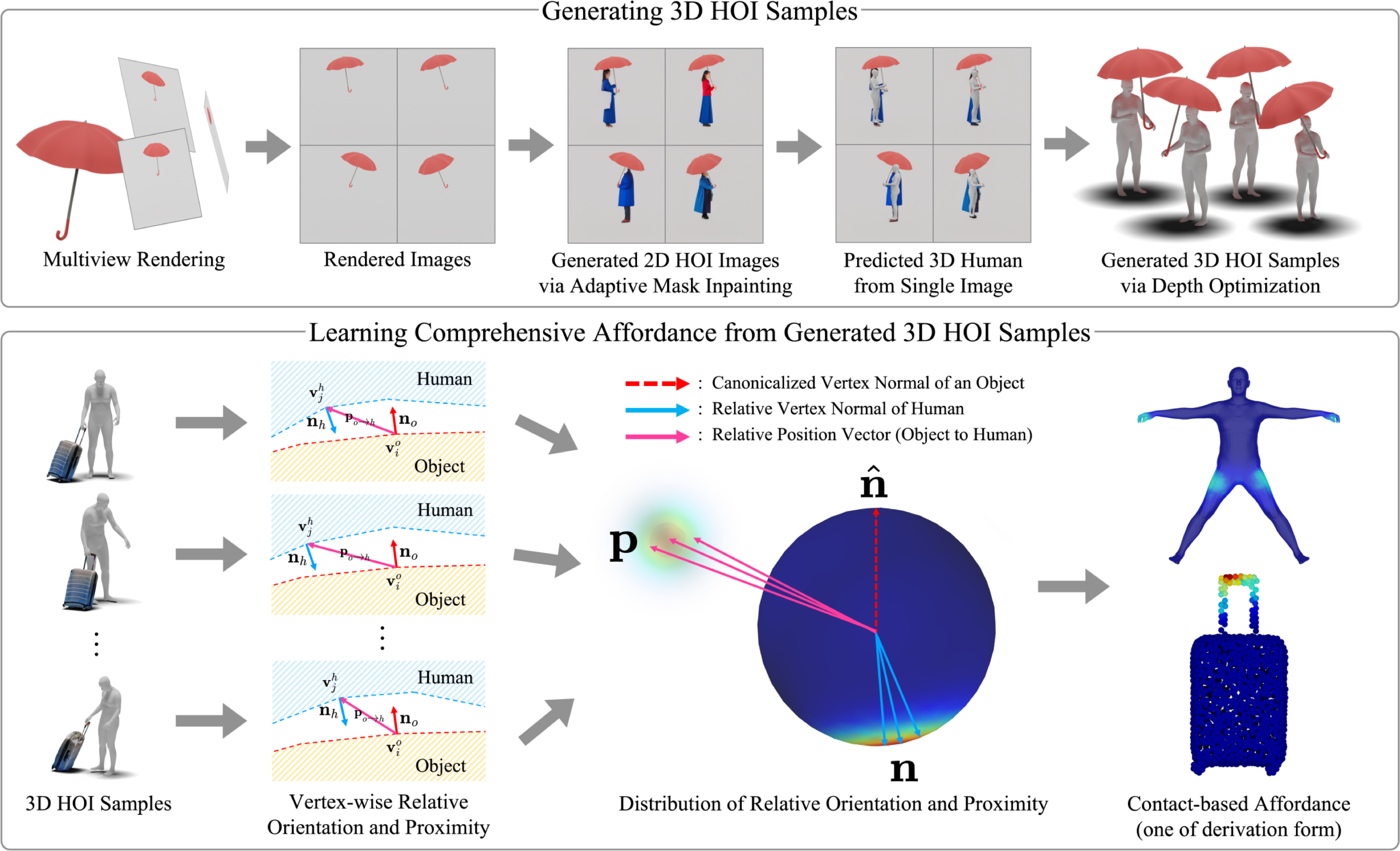}
\vspace{-15pt}
\captionof{figure}{\textbf{Method Overview.} 
Our method can be divided into two parts: (1) Generating 3D HOI Samples and (2) Learning ComA from Generated 3D HOI Samples. In the first step, we utilize an inpainting diffusion model with our \textit{Adaptive Mask Inpainting} to create 2D HOI images, and generate 3D HOI samples via uplifting pipeline. In the second step, the generated 3D HOI samples are aggregated to create distributions for relative proximity and orientation, which can be derived into various affordance forms.
}
\label{fig:overview}
\vspace{-20pt}
\end{figure}

Given an input 3D object mesh, our goal is to learn the novel affordance representation, ComA, which models the distribution of relative proximity and orientation between human and the object surface points.
Our pipeline begins with creating 2D HOI images (Sec.~\ref{subsec:affordancegeneration}). These images are then lifted to 3D, resulting large-scale synthetic 3D HOI samples (Sec.~\ref{subsec:lifting2daffordanceto3d}). From these 3D samples, we finally learn ComA (Sec.~\ref{subsec:comprehensiveaffordance}). An overview of our method is shown in Fig.~\ref{fig:overview}.

\subsection{2D HOI Image Generation}
\label{subsec:affordancegeneration}

\noindent \textbf{Rendering Object from Multi-Viewpoints.} 
Given a 3D object, we first render it from multiple viewpoints. We place the object and camera differently for two object types: \textit{static} and \textit{dynamic}. For objects that are presumed to be fixed on the ground during interaction (\textit{static}: \emph{e.g.}, chair), we fix the object on the ground plane, whereas objects with dynamic pose during interaction (\textit{dynamic}: \emph{e.g.}, cup) are perturbed with sampled rotation and translation. Type of the object is inferred automatically by rendering the object at any direction first and then utilizing vision-language model~\cite{gpt4v} for predicting. The rendering process can be fully automated, but we empirically find it beneficial to leverage a small amount of human labor, such as restricting the range of perturbation due to the inherent bias of pre-trained diffusion models. For rendering, we place weak perspective cameras $\{\Pi_c\}_{c=1}^N$ around the object with equal azimuth interval and same elevation, where $N$ is set as 8 for \textit{static}, 40 for \textit{dynamic} objects.

\noindent \textbf{Inpainting Mask Selection.} 
After rendering the images, we select the masks to inpaint a human interacting with the object. While our inpainting pipeline is quite robust to initial masks, we found that initial mask selection is helps avoid failure cases such as generating hallucinated objects or ambiguous interactions. Specifically, we move sliding windows around the object and choose the windows whose Intersection over Union (IoU) with the object segmentation mask falls within a specific range. See Supp. Mat. for more details.

\noindent \textbf{Prompt Generation.}
Inspired by CHORUS~\cite{CHORUS}, we automatically generate the prompts that describe the possible interactions with humans and the given object, which are taken as guidance for inpainting. Specifically, we utilize the vision-language model~\cite{gpt4v} to infer the HOI prompt using the rendered object image and predefined template. We generate 3 prompts per given object.

\begin{figure}[t]
\centering
\includegraphics[width=\columnwidth]{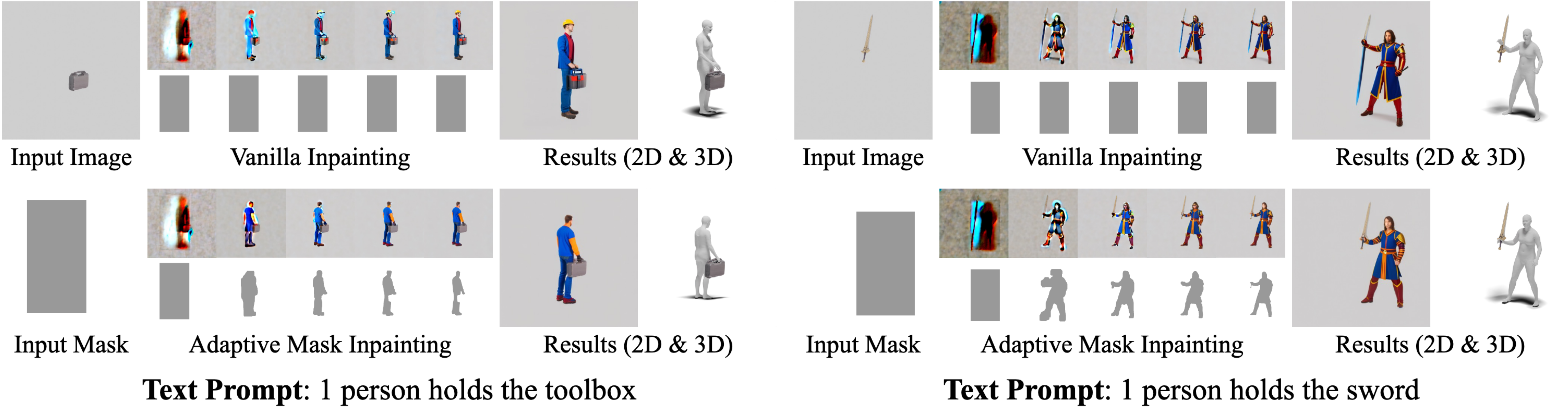}
\vspace{-15pt}
\caption{\textbf{Adaptive Mask Inpainting.} Without Adaptive Mask Inpainting, the original object is damaged when inserting humans, resulting in false affordances.}
\label{fig:adaptivemaskinpainting}
\vspace{-15pt}
\end{figure}

\noindent \textbf{Adaptive Mask Inpainting for Human Insertion.} We synthesize 2D images of human-object interaction by inpainting the human into rendered object images using publicly available text-conditional inpainting diffusion models~\cite{Rombach_2022_CVPR}. However, the challenge arises as the object geometry and details within the mask region are not preserved during inpainting, resulting in false affordances. To mitigate this problem, we present Adaptive Mask Inpainting to progressively specify the inpainting region over diffusion timesteps. Let the sequence of denoised image latent be $\{x_t\}_{t=T}^0$, where $x_T$ is the fully random noise and $x_0$ is the denoised latent. Inspired by the observation that the quality of the predicted denoised image $\hat{x}_0$ improves over progress of timestep, and creates a low-level structure of the target prompt (in this case, human) even at the early steps as shown in Fig.~\ref{fig:adaptivemaskinpainting}, we aim to \textit{adapt} the inpainting region over timesteps by spatially discovering the mask from the low-level structure. Specifically, we apply PointRend~\cite{pointrend} to decoded latent $\mathbf{D}(\hat{x}_0)$ at specific timesteps to predict the human mask, and consequently use as the mask for the next denoising step.

\subsection{Lifting 2D Affordance to 3D}
\label{subsec:lifting2daffordanceto3d}

From the previous section, we obtain a set of 2D HOI images $\{I_{d}\}_{d=1}^{D}$.
Even though we have multiple views with geometrically consistent object renderings, the inserted humans have diversity without multi-view consistency, and thus it is non-trivial to directly lift 3D humans from these 2D images.

To address this, we first utilize a single view 3D human prediction model to obtain a corresponding 3D human pose and shape of the image, and then estimate the depth of the human with weak auxiliary cue provided by the set of 2D HOI images, generating a 3D HOI sample corresponding to the image.

\noindent \textbf{Single View 3D Human Prediction.} To uplift the 2D HOI images into 3D, we first predict the pose and shape of 3D humans from the images. We apply off-the-shelf 3D human prediction model~\cite{Moon_2022_CVPRW_Hand4Whole} (denoted as $\mathbf{F}_{\text{human}}$) to predict SMPL-X humans~\cite{SMPL-X:2019} from generated images:
\begin{equation}
    \{\mat{\theta}_d, \vec{\beta}_d, \mathbf{j}_d, \vec{R}_d^h, \mathbf{s}_d^h, x_d, y_d\}= \mathbf{F}_{\text{human}}(I_d)
    \label{eq:fhuman}
\end{equation}
where $\mat{\theta}_d\in\mathbb{R}^{54\times3}, \vec{\beta}_d\in\mathbb{R}^{10}, \mathbf{j}_d\in\mathbb{R}^{65\times3}$ are predicted SMPL-X pose, shape, and joints (see Supp. Mat. for the increased number of joints), respectively. $\vec{R}_d^h\in\text{SO}(3)$ is global rotation, and $\mathbf{s}_d^h, x_d, y_d$ are each scale, and $x,y$ direction offsets representing the parameters of weak perspective camera $\Pi_{\gamma(d)}$ assigned to image $I_d$. $\gamma(\cdot)$ is the function that maps the index of the 2D HOI image to the index of the weak perspective camera from which it is rendered.
We additionally define $\mathbf{j}_{d^{\dagger}}$ as joints in world coordinate, obtained by transforming inferred joints $\mathbf{j}_d$ with the camera rotation and translation of $\Pi_{\gamma(d)}$. As 3D human lies within the weak perspective camera framework, there remains depth ambiguity to solve.

\noindent \textbf{Depth Optimization using Weak Auxiliary Cue.} To address the depth ambiguity of 3D human obtained from reference image $I_d$, we leverage 2D HOI images generated in advance. Since the human poses in images are geometrically inconsistent, we select the largest subset inliers $\mathcal{I}$ among them which are ``semi-consistent'' with the reference image by applying RANSAC~\cite{fischler1981random} in terms of joint re-projection error. The joint triangulation is performed on two images, including the reference image. Note that we generate sufficiently large amount of 2D HOI images to ensure the existence of the inlier set. The inliers $\mathcal{I}$ is used to estimate the human depth $z_d$ by optimizing the joint re-projection loss defined below:

\begin{equation}
    \mathcal{L}_{\text{re-projection}} = {1\over |\mathcal{I}|}\sum_{I_{\delta} \in \mathcal{I}} \| {\Pi_{\gamma(\delta)}}(\mathbf{j}_{d^{\dagger}} + z_d\bm{f}_d) - {\Pi_{\gamma(\delta)}}(\mathbf{j}_{\delta^{\dagger}}) \|^2
\end{equation}
where $\bm{f}_d\in\mathbb{R}^{3}$ is the normalized direction of the camera $\Pi_{\gamma(d)}$. 
Inspired by Jiang \etal~\cite{Jiang_2020_CVPR}, we leverage the human segmentation masks to reason about occlusion and promote a initial depth for human.
We initialize SMPL-X~\cite{SMPL-X:2019} models along the camera direction and select the model with the best IoU between the rendered mask (regarding occlusion) and the segmented region to be an initial of optimization.
This is useful when the optimal object position is in between two human body parts along the camera direction (\emph{e.g.}, a motorcycle body between the legs while riding) as we use a collision loss term during optimization.

The collision loss is used to estimate fine-grained depth, as the collision cue can provide physical plausibility of 3D HOI samples. We adopt collision loss from COAP~\cite{Mihajlovic:CVPR:2022_COAP} where the human body is represented as signed distance field~\cite{oleynikova2016signeddistancefields}. We set object vertices as query points to reduce collision between human and object, namely $\mathcal{L}_\text{collision}$. We optimize depth $z_d$ to minimize the total loss:
\begin{equation}
    \mathcal{L} = \mathcal{L}_{\text{re-projection}} + \lambda_{\text{collision}}\mathcal{L}_{\text{collision}}
\end{equation}

\noindent\textbf{Filtering.} We filter out potentially low-quality 3D human samples in the cases of: (1) human rendering (regarding occlusion) and predicted human segmentation do not overlap much; (2) number of inliers during RANSAC~\cite{fischler1981random} is small; (3) human and object collides significantly in 3D.

\subsection{Learning Comprehensive Affordance}
\label{subsec:comprehensiveaffordance}

Our main objective is to learn ComA, a new affordance representation which encompasses orientational and spatial affordance, along with contact-based affordance. In order to learn ComA from the previously generated 3D HOI samples, we first apply Poisson disk sampling~\cite{poissondisk2} to uniformly select 3D surface points from the object and uniformly sample mesh vertices from SMPL-X~\cite{SMPL-X:2019} human, constructing ordered point sets for each object and human.

Given the ordered point sets, the density function on the $i$-th object point and the $j$-th human point, which we denote as $\mathcal{P}_{ij}(\mathbf{p},\mathbf{n})$, represents the joint probability of the relative position $\mathbf{p}\in\mathbb{R}^3$ and relative normal orientation $\mathbf{n}\in\mathbb{S}^2$ of the $j$-th human point with respect to $i$-th object point within the 3D HOI samples.
We obtain $\mathbf{p}$ and $\mathbf{n}$ by canonicalizing the orientation of the human surface normal $\mathbf{n}^h_j$ and relative position $\mathbf{p}^{o\xrightarrow[]{}h}$, assuming the object surface normal $\mathbf{n}^o_i$ is rotated to face $\mathbf{\hat{n}} = [0,0,1]^T$ (see Supp. Mat. for details on canonicalization).
The distribution $\mathcal{P}_{ij}$ sums up to 1 for all possible $\mathbf{p}$ and $\mathbf{n}$:
\begin{equation}
    \int_{\mathbb{R}^3}\int_{\mathbb{S}^2} \mathcal{P}_{ij}(\mathbf{p},\mathbf{n})d\mathbf{n}d\mathbf{p} = 1
\end{equation}
In practice, we set the domain of $\mathbf{p}$ as voxelized grid and domain of $\mathbf{n}$ as equispaced grid on $\mathbb{S}^2$, obtained via Fibonacci spirals~\cite{fibonaccispiral}, and compute the discrete probabilities by aggregating Gaussian kernels calculated for $\mathbf{n}$ and $\mathbf{p}$ (we use geodesic metrics for $\mathbf{n}$).
Under the probability distribution $\mathcal{P}_{ij}(\mathbf{p}, \mathbf{n})$, we define three types of functions $f(\mathbf{p},\mathbf{n})$ capturing different aspects of affordances: (1) Contact-based Affordance, (2) Orientational Affordance, (3) Spatial Affordance. Given each function $f(\mathbf{p}, \mathbf{n})$, the pointwise affordance between $i$-th object point and $j$-th human point is determined by computing its expectation:
\begin{equation}
    \mathbb{E}_{\mathbf{p}, \mathbf{n}\sim \mathcal{P}_{ij}}\,[f(\mathbf{p}, \mathbf{n})]
    \label{affordance_exp_eq}
\end{equation}

\noindent \textbf{Contact-based Affordance.}
From ComA, we can infer the contact score for human-object point pairs regarding proximity and normal alignment as:
\begin{equation}
f_\text{contact}(\mathbf{p},\mathbf{n}) = {\left( 1-\mathbf{n}\cdot \mathbf{\hat{n}}\over 2 \right)}e^{-\|\mathbf{p}\|}
\label{eq:contactaffordance}
\end{equation}
where $e^{-\|\mathbf{p}\|}$ term encourages high score when distance $\|\mathbf{p}\|$ is close. The ${1-\mathbf{n}\cdot \mathbf{\hat{n}}\over 2}$ term, motivated by the fact that physical interactions are conducted via exerted normal force between contact points, encourages high score when the object normal $\mathbf{n}_i^o$ and human normal $\mathbf{n}_j^h$ align in an antiparallel direction. The normal alignment term improves the precision of the contact map, especially when small body parts 
(\emph{e.g.}, fingertips) interact with an object and allows robust affordance learning even when using noisy HOI datasets, compared to prior work that only uses proximity for contact terms~\cite{BEHAVE, POSA, yi2022mover, PSI, prox} or normal for computing only penetration~\cite{grady2021contactopt, yang2021cpf}, without considering the alignment of surface normals.

\noindent \textbf{Orientational Affordance.} 
We aim to capture the pattern and tendency of the orientation of body parts with respect to the object based on the concept of Shannon entropy~\cite{shannon2001mathematical}. Intuitively, lower entropy means there exists more typical patterns of human body orientations during the interactions.
To quantify the orientational tendency for ComA, we use negated normalized Shannon entropy~\cite{shannon2001mathematical} to enforce high value when the human surface normal shows consistent orientational tendency (low variance) with respect to the object surface normal:
\begin{equation}
    \label{eq:implicitresponse}
    f_\text{orientation}(\mathbf{n}) = 1+{\log \mathcal{P}_{ij}(\mathbf{n})\over \log n_b}
\end{equation}
where $n_b$ denotes the number of discretized bins of probability measure. 
Intuitively, the orientational affordance defined by $\mathbb{E}_{\mathbf{n}\sim \mathcal{P}_{ij}}\,[f_\text{orientation}(\mathbf{n})]$ has a low value when marginalized distribution $\mathcal{P}_{ij}(\mathbf{n})$ is close to the uniform distribution, meaning that there is no dominant orientational pattern.
Note that Eq.~\ref{eq:implicitresponse} is agnostic to the proximity term $\mathbf{p}$, which allows us to model the nonphysical orientational effects even for far-distance points.

\noindent \textbf{Spatial Affordance}. Following CHORUS~\cite{CHORUS}, we also consider the 3D spatial occupancy distribution of a human surface point with respect to the object point by simply counting the occurrence:
\begin{equation}
    f_\text{spatial}(\mathbf{p})=\boldone(\mathbf{x}-\mathbf{p})
\end{equation}
where $\boldone(\cdot): \mathbb{R}^3\xrightarrow[]{}\{0,1\}$ is a binary function that returns 1 if the argument is zero vector, else 0. Thus the spatial affordance $\mathbb{E}_{\mathbf{p}\sim \mathcal{P}_{ij}}\,[f_\text{spatial}(\mathbf{p})]$ outputs scalar occupancy for the spatial vector 
$\mathbf{x}\in\mathbb{R}^3$. In practice, we implement the function as voxel array, counting 1 to the voxel that contains $\mathbf{p}$. Note that $f_\text{spatial}$ is agnostic to the normal direction $\mathbf{n}$ via marginalization. Learning the occupancy informs us about macro positioning of the human relative to the given object.
\section{Experiments}
\label{sec:experiments}

In this section, we conduct experiments to demonstrate the efficacy of our representation ComA, and the method for learning it. In Sec.~\ref{subsec:qual}, we illustrate how ComA extends beyond contact-based affordance to learn distributions for orientational tendency and spatial relation, by visualizing it across various object categories. In Sec.~\ref{subsec:quant}, we perform quantitative comparisons with other baselines and verify the efficacy of our Adaptive Mask Inpainting.
Finally, in Sec.~\ref{subsec:application}, we introduce an optimization framework achievable through the knowledge provided by ComA, demonstrating the potential of our new affordance representation.

\subsection{Datasets}
\label{subsec:datasets}

\noindent \textbf{3D Object Datasets.} 
We obtain 3D object meshes from various 3D object datasets for qualitative evaluation. Specifically, We utilize 20 from BEHAVE~\cite{BEHAVE}, 10 from InterCap~\cite{InterCap}, 2 from ShapeNet~\cite{ShapeNet}, and 8 from SAPIEN~\cite{SAPIEN} to learn ComA individually. As textures were not available in InterCap~\cite{InterCap}, making the diffusion model difficult to inpaint, we use TEXTure~\cite{TexTure} to generate textures on the meshes using text prompt automatically generated from ChatGPT~\cite{chatgpt}.

For quantitative evaluation, we utilize BEHAVE~\cite{BEHAVE} as a ground truth dataset to verify the quality of generated 3D HOI samples, assuming that interactions captured in lab-controlled environments exhibit the upper-bound quality of HOI. BEHAVE~\cite{BEHAVE} contains video sequences of HOI scenarios for 20 object categories.
Each video frame is annotated with a single 3D human and object, where each human is given as SMPL-H~\cite{MANO:SIGGRAPHASIA:2017,SMPL:2015} format and object as rotation and translation applied to the canonical mesh. 
We preprocess the dataset by transferring the SMPL-H annotations into SMPL-X format following Choutas \etal~\cite{SMPL-X:2019}. We use the 3D human-object pair in each frame as a sample, compute vertex-wise contact scores in the same way as our method, and use them as ground truth.

\noindent \textbf{Internet Search Datasets.} 
We also collect 60 free 3D object meshes from Internet source, SketchFab~\cite{SketchFab} and learn ComA.

\subsection{Baselines and Metric}
\label{subsec:baseline_metric}

As previous studies focused on contact rather than learning the overall aspect of affordances, we use contact-based affordance derived from ComA to compare with them. However, most of them output 3D contact of a single HOI situation by inferring affordance from a given HOI sample (mostly an image), while we produce an overall affordance distribution for the given 3D object. To address this, we treat the contact inferred by state-of-the-art methods, DECO~\cite{DECO} (for the human) and IAGNet~\cite{IAGNet} (for the object) from the HOI image as a single sample, and aggregate them across multiple images to construct a distribution for comparison against BEHAVE~\cite{BEHAVE}.

Specifically, we aggregate the results of DECO~\cite{DECO} and IAGNet~\cite{IAGNet} on (1) 2D HOI images we've generated beforehand and (2) BEHAVE~\cite{BEHAVE} test images, as these represent the usage of the BEHAVE~\cite{BEHAVE} object. 
The aggregated scores are then normalized to form a distribution (summing up to 1), considering only the relative probability of potential contact, as the original scale of the contact scores varies between methods. Subsequently, we compare similarity scores (SIM)~\cite{swain1991color} and mean absolute error (MAE)~\cite{MAE} between these distributions.

\begin{figure}[t]\centering
\includegraphics[width=0.96\linewidth, trim={0 0 0 0},clip]{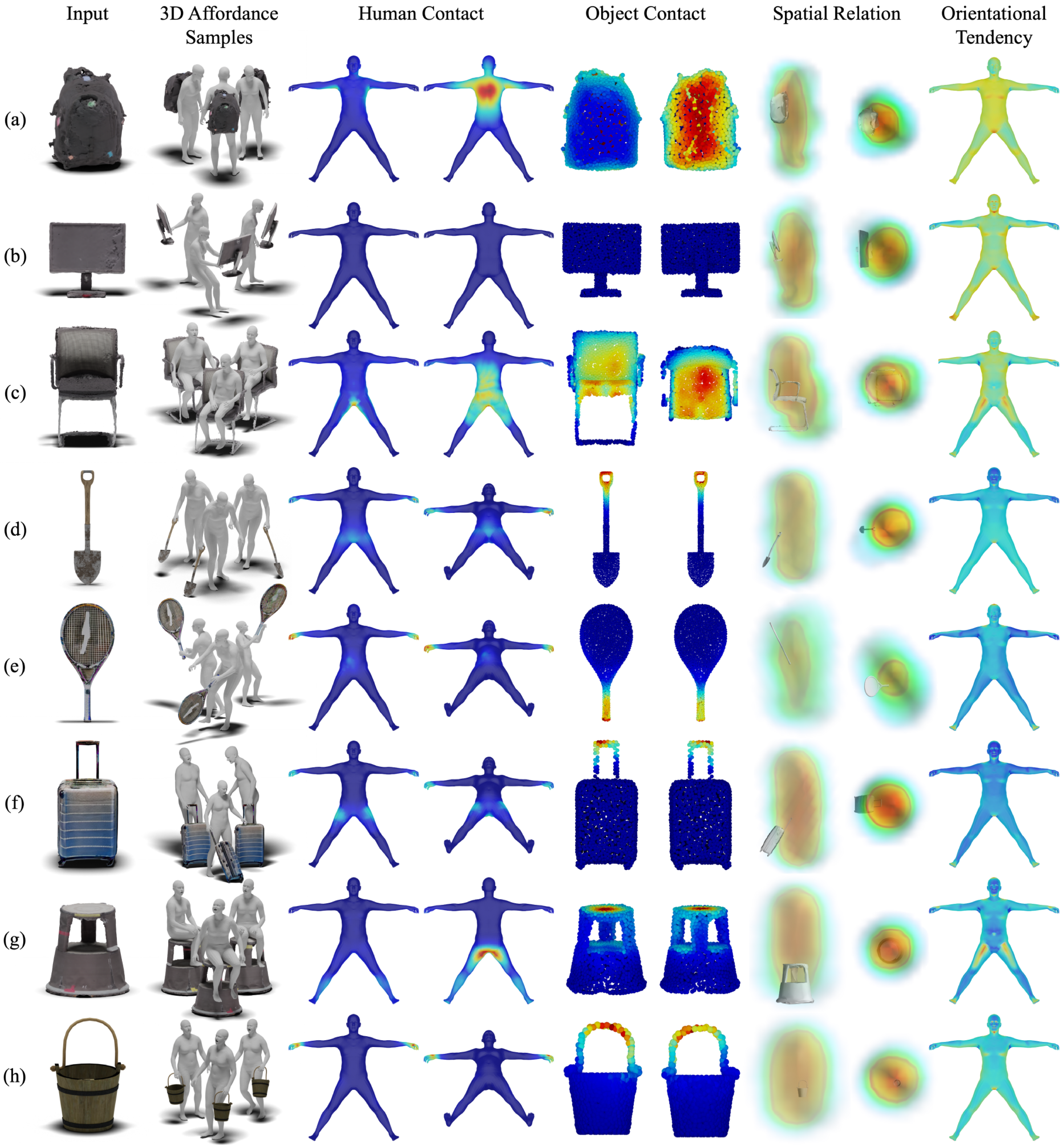}
\vspace{-7pt}
\captionof{figure}{\textbf{Qualitative Results.}
ComA can model distributions of contact, orientation, and spatial relation exhibited during the interaction between humans and novel objects.
}
\label{fig:main_qual}
\vspace{-25pt}
\end{figure}

\subsection{Qualitative Results}
\label{subsec:qual}

\noindent \textbf{Contact for Various Objects.} 
Similar to other studies, ComA can be derived into contact-based affordance as shown in Fig.~\ref{fig:main_qual}. As our representation provides contact scores for all pairs of human-object vertices, we assign the contact score of each vertex as the maximum score among the pairs that include it. We can observe the conventional object usage patterns as heatmap, which fits with general human usages. Specifically, in case Fig.~\ref{fig:main_qual}(b), where significant contact is absent, we can infer that the interaction does not frequently appear by contact.

\noindent \textbf{Orientational Tendency for Various Objects.} 
ComA also contains information on orientational tendencies during the HOI. The last column in Fig.~\ref{fig:main_qual} visualizes the negated entropy of the vertex normal direction of human relative to the object as a heatmap. Fig.~\ref{fig:main_qual}(a), Fig.~\ref{fig:main_qual}(b), and Fig.~\ref{fig:main_qual}(c) exhibit relatively high orientational tendency, while the rest show lower.
This aligns with the fact that the human's orientation is fixed by the attachment to the back as in Fig.~\ref{fig:main_qual}(a), and by the restrictions present on the back and side of the object as in Fig.~\ref{fig:main_qual}(c).
In Fig.~\ref{fig:main_qual}(b), there exists an orientational tendency without any attachments or restrictions, which arises due to long-distance interaction: viewing.

\noindent \textbf{Spatial Relation for Various Objects.} 
The relative position of the human during HOI can be also derived from ComA. Fig.~\ref{fig:main_qual} visualizes the distribution of the human's full body positions relative to the object.
Specifically, as shown in Fig.~\ref{fig:main_qual}(d), Fig.~\ref{fig:main_qual}(e), and Fig.~\ref{fig:main_qual}(f), we find that the spatial relations for the object with multiple utilities and geometries (\emph{e.g.}, in the case of shovel, handle for holding and blade for digging) align with general human usage, demonstrating the plausibility of the results.
Especially, in the case of tennis racket shown in Fig.~\ref{fig:main_qual}(e), we can observe a relatively weak tendency in spatial relation, as the racket can move almost freely relative to human during the interaction.

\begin{figure}[t]\centering
\includegraphics[width=0.9\linewidth, trim={0 0 0 0},clip]{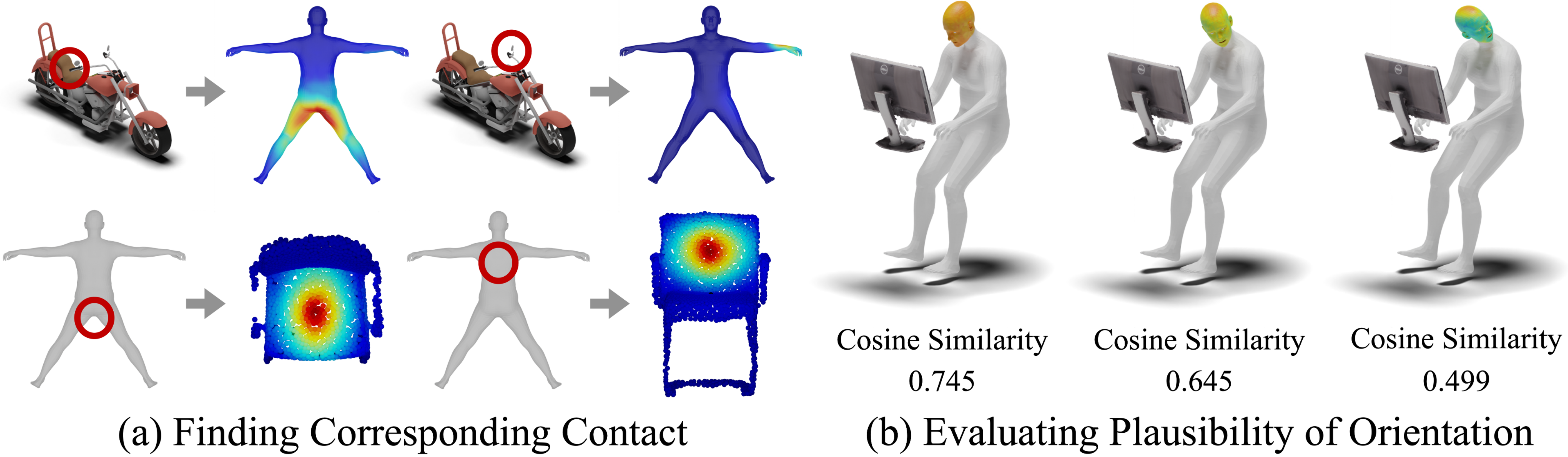}
\vspace{-5pt}
\captionof{figure}{\textbf{Pointwise Affordance Knowledge.} 
ComA can provide pointwise affordance knowledge such as corresponding contacts or plausible orientation of body parts.
}
\label{fig:prior_knowledge}
\vspace{-5pt}
\end{figure}

\noindent \textbf{Pointwise Affordance Knowledge.} 
Different from previous studies, ComA models the distribution of relative orientation and proximity for all human-object vertex pairs, allowing it to provide pointwise affordance knowledge such as corresponding contacts and plausible orientation direction of specific body parts.
As shown in Fig.~\ref{fig:prior_knowledge}(a), we can infer how each human parts and object parts contact. For the case of motorcycle, we can see that saddle and handle each contacts with human's hips and hands; while for the case of chair, seat and backrest each contacts with human's hips and back.
Also, we can utilize ComA to infer the probable direction of the head while using the monitor by computing the cosine similarity between the head direction and the vertex normal with highest orientational tendency. As illustrated in Fig.~\ref{fig:prior_knowledge}(b), we can see that the similarity maximizes when gazing towards the screen; which is a plausible head direction when interacting with the monitor. 

\noindent \textbf{Ablation on Adaptive Mask Inpainting.}
As shown in Fig.~\ref{fig:adaptivemaskinpainting}, without Adaptive Mask Inpainting, the original geometry of the object changes (left) or gets replaced by hallucinated objects (right), resulting in false affordances unfaithful to the original object. 
Consequently, the difference between the object in 2D and the original one existing in 3D causes the uplifted 3D HOI samples to fail to represent the right usage.
This shows the importance of Adaptive Mask Inpainting as it prevents 2D inpainting diffusion model from generating false affordances.

\begin{table}[t]
    \small
    \centering
    \resizebox{0.95\columnwidth}{!}{
        \begin{tabular}{lcccc}
        \toprule
        Methods & SIM$_\text{Human}$(\%)$\uparrow$ & SIM$_\text{Object}$(\%)$\uparrow$ & MAE$_\text{Human}$($\times 100$)$\downarrow$ & MAE$_\text{Object}$($\times 100$) $\downarrow$ \\
        \midrule \midrule
        Ours & \textbf{52.82} & \textbf{70.55} & \textbf{0.126} & \textbf{0.0288} \\
        IAGNet~\cite{IAGNet}$_\text{ on Our 2D HOI Images}$ & - & 66.34 & - & 0.0329 \\
        IAGNet~\cite{IAGNet}$_\text{ on BEHAVE Test Images}$ & - & 63.86 & - & 0.0353 \\
        DECO~\cite{DECO}$_\text{ on Our 2D HOI Images}$ & 21.66 & - & 0.246 & - \\
        DECO~\cite{DECO}$_\text{ on BEHAVE Test Images}$ & 17.39 & - & 0.259 & - \\
        \bottomrule
        \end{tabular}
    }
\vspace{5pt}
\caption{\textbf{Quantitative Results on BEHAVE.} We report SIM, MAE to evaluate the similarity of contact distribution (normalized contact scores) on BEHAVE sequences.}
\label{tab:quant}
\vspace{-28pt}
\end{table}

\begin{table}[t]
    \scriptsize
    \centering
    \resizebox{0.65\columnwidth}{!}{
        \begin{tabular}{lccc}
        \toprule
        {Inpainting Methods}        &RMSE$_\text{Background}\downarrow$&mIoU$\uparrow$&mIoU$_\text{Occlusion Aware}\uparrow$ \\
        \midrule \midrule
        Vanilla & 42.23 & 0.535 & 0.631 \\
        Adaptive Mask &\textbf{15.33} & \textbf{0.706} & \textbf{0.815} \\
        \bottomrule
        \end{tabular}
    }
\vspace{5pt}
\caption{\textbf{Effects of Adaptive Mask Inpainting} We evaluate the performance of background preservation in vanilla inpainting and our Adaptive Mask Inpainting.}
\label{tab:adaptivemask_preservation}
\vspace{-27pt}
\end{table}

\subsection{Quantitative Results}
\label{subsec:quant}

\noindent \textbf{Comparison with Baselines.} 
We evaluate the similarity of contact distribution between BEHAVE~\cite{BEHAVE} and each of DECO~\cite{DECO}, IAGNet~\cite{IAGNet}, and ours.
We sample 2048 points from the 3D object mesh surface following IAGNet~\cite{IAGNet} and only consider the contact underneath the head for DECO~\cite{DECO}, as DECO~\cite{DECO} uses SMPL~\cite{SMPL:2015} format, which only has vertex correspondence with SMPL-X~\cite{SMPL-X:2019} in non-head regions.
As shown in Tab.~\ref{tab:quant}, the contact distribution from IAGNet~\cite{IAGNet} and DECO~\cite{DECO} using our 2D HOI images outperforms the one modeled from BEHAVE test images.
This means that our generated HOI images represent the real-world affordance as well as BEHAVE~\cite{BEHAVE} test images, demonstrating the efficacy of leveraging pre-trained diffusion model for generating affordance.
Additionally, the contact distribution from our method outperforms the one modeled using IAGNet~\cite{IAGNet} and DECO~\cite{DECO} with our generated HOI images. Since we extract 3D contacts from the same image sets, we argue that our pipeline is better than other models trained on annotations for uplifting 2D affordance to 3D.

\noindent \textbf{Efficacy of Adaptive Mask Inpainting.} 
To verify the effectiveness of our Adaptive Mask Inpainting, we compare the amount of background preservation for our generated 2D HOI images with the vanilla inpainting diffusion model.
We sample at most 2 inpainted images with a single human detected, per each object renderings for all categories in BEHAVE~\cite{BEHAVE}; resulting in 691 images for vanilla inpainting and 692 images for Adaptive Mask Inpainting.
We compute 3 metrics for these images; (1) $\text{RMSE}_{\text{Background}}$: pixel error between images excluding the predicted human region; (2) $\text{mIoU}$: average IoU between object predictions in each inpainted image and original object image; (3) $\text{mIoU}_{\text{Occlusion Aware}}$: same metric as (2) but excluding the predicted human region during computation. To ensure fairness, we use open vocabulary segmentation model~\cite{liu2023groundingdino, Kirillov_2023_ICCV_SegmentAnything} to predict detailed human/object segmentation.
As shown in Tab.~\ref{tab:adaptivemask_preservation}, we demonstrate that our adaptive mask algorithm better preserves the background, not only pixelwise (RMSE$_\text{Background}$), but also semantic-region-wise (mIOU, mIOU$_\text{Occlusion Aware}$) than the vanilla inpainting baseline. This means that the generated images from Adaptive Mask Inpainting are more likely to preserve the original object.

\subsection{Application}
\label{subsec:application}
\begin{figure}[t]\centering
\includegraphics[width=0.85\linewidth, trim={0 0 0 0},clip]{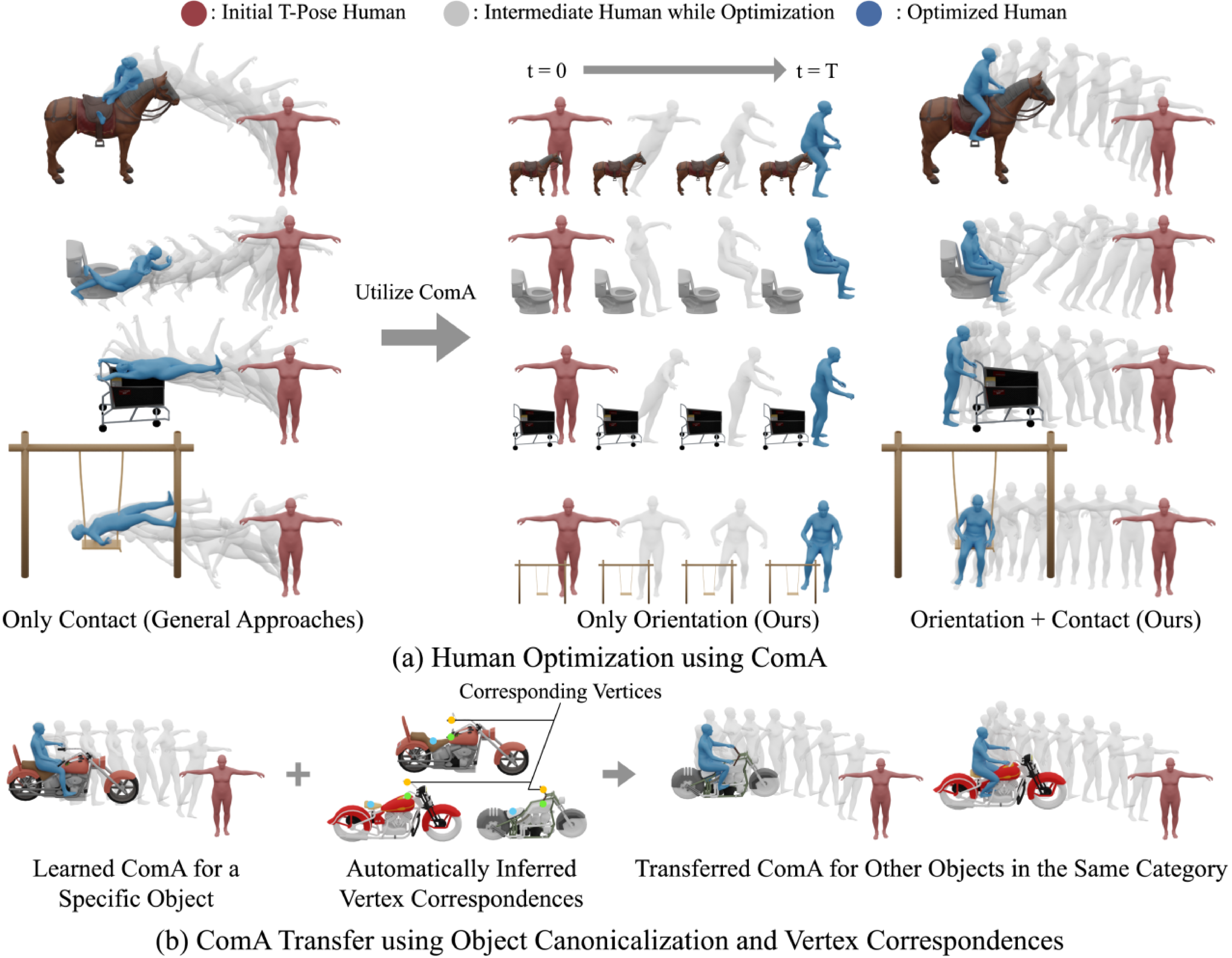}
\vspace{-5pt}
\captionof{figure}{\textbf{Application.}
ComA can reconstruct HOI through optimization, which can be scaled to other objects in same category by transferring such knowledge between them.
}
\label{fig:application}
\vspace{-17pt}
\end{figure}

As ComA is a newly proposed affordance representation, we demonstrate its potential by showcasing an application.
One possible task is to reconstruct HOI by optimizing the human pose and position to interact with an object.
As shown in Fig.~\ref{fig:application}(a), it is nearly impossible to achieve the task with traditional methods which relies on contact affordance only.
In contrast, ComA provides both relative orientation and contact during the HOI, making it possible to generate plausible 3D HOI samples from T-posed humans. For the objects with multiple modes of affordances, we apply mode seeking algorithm~\cite{MeanShift} on 3D HOI samples with their vertex normals and learn ComA separately; for example, swing in Fig.~\ref{fig:application}(a).

However, ComA learned on a specific object cannot be directly applied to other objects with different geometries. To address this, we transfer ComA through the process of (1) object canonicalization and (2) finding vertex correspondences. Fig.~\ref{fig:application}(b) provides an example of transferring ComA learned from a specific motorcycle to other motorcycles with different geometries. We use objects from ShapeNet~\cite{ShapeNet} which are already canonicalized and apply CPNet~\cite{CPNet} to find vertex correspondences automatically for transferring the vertices with contact. As object canonicalization and vertex correspondence can each transfer macro orientation information and local contact information, we can transfer ComA and scale the human optimization task to other objects in the same category without any additional learning.

\section{Conclusion}
\label{sec:conclusion}

In this paper, we point out the importance of learning non-contact patterns in HOIs and propose a novel affordance representation called ComA which models both contact and non-contact patterns. To learn ComA, we utilize a pre-trained 2D diffusion model which already possesses prior knowledge of affordance. To leverage the affordance knowledge, we design a Rendering-Inpainting-Uplifting pipeline to generate 3D HOI samples and learn ComA from the samples. The results show that ComA effectively models the common usage of objects in terms of contact, orientation, and spatial relations.
Within the pipeline, we propose \textit{Adaptive Mask Inpainting}, which allows insertion of human without damaging the original image, where we evaluate its efficacy both qualitatively and quantitatively.
We expect ComA to be used as a prior knowledge for various downstream tasks, where we demonstrate its potential through human optimization framework and the possibility of transfer to various objects in the same category.

\section*{Acknowledgements}
This work was supported by Naver Webtoon. The work of SNU members was also supported by NRF grant funded by the Korean government (MSIT) (No.
2022R1A2C2092724 and No. RS-2023-00218601), and IITP grant funded by the Korean government (MSIT) [RS-2021-II211343, AI Graduate School Program (SNU)]. H. Joo is the corresponding author.

\bibliographystyle{splncs04}
\bibliography{egbib}

\title{Supplementary Materials for \\Beyond the Contact: Discovering Comprehensive Affordance for 3D Objects from Pre-trained 2D Diffusion Models}

\titlerunning{Discovering Comprehensive Affordance for 3D Objects}

\author{Hyeonwoo Kim\inst{1*} \and
Sookwan Han\inst{1*} \and
Patrick Kwon\inst{2} \and
Hanbyul Joo\inst{1}
}

\authorrunning{H. Kim et al.}

\institute{Seoul National University \\
\and
Naver Webtoon AI\\
}
\maketitle
\def\thefootnote{*}\footnotetext{Indicates equal contribution}\def\thefootnote{\arabic{footnote}}

\renewcommand*{\thesection}{\Alph{section}} 
\renewcommand{\theequation}{S.\arabic{equation}} 
\renewcommand{\thefigure}{S.\arabic{figure}} 
\renewcommand{\thealgorithm}{S.\arabic{algorithm}}

\section{Implementation Details}

We provide further details on the implementations of our pipeline (Sec.~\ref{subsec:affordancegeneration}$\sim$\ref{subsec:lifting2daffordanceto3d}) and the formulation of ComA (Sec.~\ref{subsec:comprehensiveaffordance}) in our main paper.

\subsection{Rendering Object from Multi-Viewpoints}
\label{subsec:rendering}
For \textit{static} type objects, we install weak perspective cameras around the object with $45\degree$ azimuth intervals, creating 8 distinct views. 
For \textit{dynamic} type objects, we install 4 weak perspective cameras with $90 \degree$ azimuth intervals. The elevation is set constant within [$0\degree$, $30\degree$] range, where the values differ by category. For dynamic objects, we perturb the object with random rotations and translations. Specifically, we uniformly sample the rotation in the form of euler angle, where yaw, pitch, and roll are uniformly sampled from a predefined range. Random translations are sampled in a similar way, where each component of 3D displacement is sampled from a predefined range. Note that we set weak perspective camera scale and additional $z$-direction displacement of camera as hyperparameters. We repeat the rendering procedures 10 times with different perturbations, resulting in 40 distinct views.

\subsection{Inpainting Mask Selection}
\label{subsec:inpaintingmaskselection}
\begin{figure}[t]
\centering
\includegraphics[width=1.0\columnwidth]{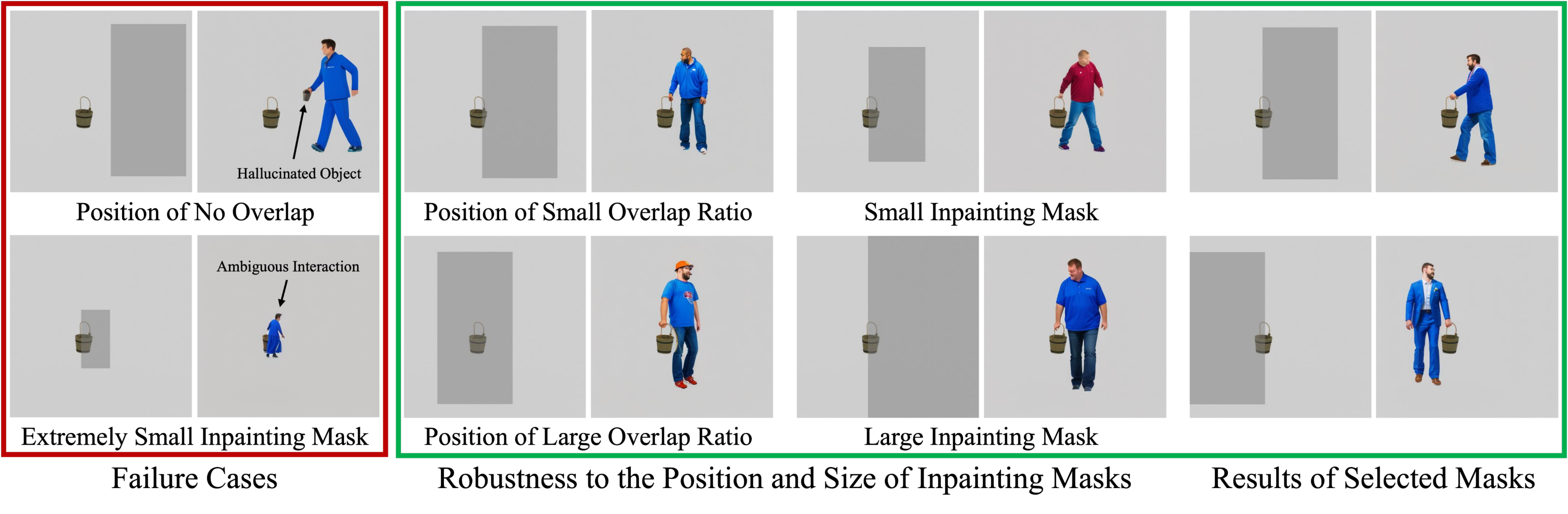}
\vspace{-15pt}
\caption{\textbf{Experiments for Mask Selection.} While our \textit{Adaptive Mask Inpainting} method is quite robust to initial masks as shown in the right (green box), our inpainting mask selection strategy helps avoid generating failure cases shown in the left (red box).
}
\label{fig:mask_selection}
\vspace{-5pt}
\end{figure}

Via thorough experiments, while the inpainting pipeline is quite robust to initial masks, we found that initial mask selection is beneficial for avoiding failure cases as shown in Fig.~\ref{fig:mask_selection}.
Specifically, we build strategies for selecting appropriate positions and sizes of masks to prevent generating: (1) hallucinated objects, which typically occurs when initial mask do not overlap with the rendered object; (2) ambiguous interactions, when inpainting masks are too small compared to the object, generating humans that ignore the relative scale with respect to the object.

The strategy starts by rendering the inpainting masks while rendering the object, using the same camera parameters. For each camera, we place an upright window perpendicular to the $xy$ plane in the world coordinate, also perpendicular to the $xy$ projection of the camera's front vector. For each mask, the center of the intersection with $z=0$ plane lies within the $xy$ projection of the 3D object. Note that the strides of the upright windows with respect to $x,y$ direction are given as hyperparameters, along with the height and width of the window. We render these upright windows using assigned cameras to obtain 2D rectangular masks, consequently used as initial inpainting masks that occlude the original object. To reduce the number of unnecessary masks (\emph{e.g.}, masks that do not cover the object but mostly the background), we only retain the masks if the Intersection over Union (IoU) between the mask and the original object lies within the predefined range, also given as hyperparameters.

\begin{figure}[t]
\centering
\includegraphics[width=1.0\columnwidth]{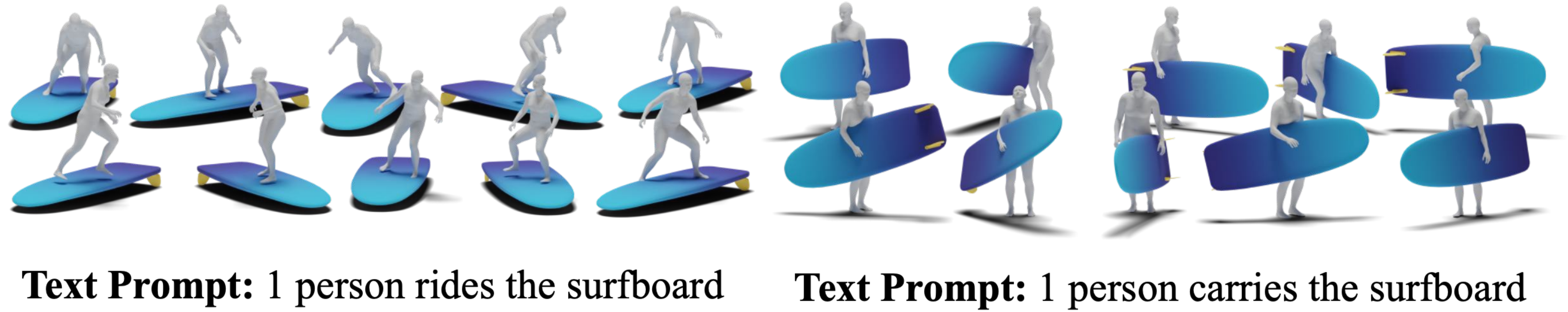}
\vspace{-5pt}
\caption{\textbf{Diversity of 3D HOI Samples and Interaction Type.} We initially create multiple HOI prompts and inpainting masks, which allows us to generate diverse 3D HOI samples with different interaction types by selecting the prompt and masks.}
\label{fig:sample_diversity}
\vspace{-15pt}
\end{figure}

\subsection{Prompt Generation} We design a generalizable pipeline to generate human-object interaction prompts even when the category of the object is unknown. We utilize GPT4v~\cite{gpt4v}, where we input the rendered object image and the following query template:

\begin{quote}
\textit{Generate at most 3 simple subject-verb-object prompt where subject’s word is exactly `1 person' and object's image is given. You should use diverse and general word but no pronoun for subject. Generated prompt must align with common sense. Verb must be simple as possible, and should depict physical interaction between subject and object. Also, only the interaction with given object is allowed, and no other objects should be introduced in the prompt.}
\end{quote}

\noindent For 3D objects of which category is already known (generally for objects obtained from SketchFab~\cite{SketchFab}), we use ChatGPT~\cite{chatgpt} to generate prompts for human-object interaction using the following query template:

\begin{quote}
\textit{Generate at most 3 simple subject-verb-object prompt where subject’s word is exactly ‘1 person’ and object's word is exactly }`\{\textbf{category}\}'. \textit{You should use diverse and general word but no pronoun for subject. Generated prompt align with common sense. Verb must be simple as possible, and should depict physical interaction between subject and object. Also, only the interaction with given object is allowed, and no other objects should be introduced in the prompt.}
\end{quote}

\noindent We do not augment the prompt except ``\textit{full body}'' at the end, where we empirically find this augmentation useful when generating the whole human body instead of a zoom-in shot of body parts (\emph{e.g.}, face, hand). The generated prompts usually describe different types of HOI, which allows our pipeline to generate diverse 3D HOI samples by altering the input prompt or varying inpainting masks in multiview renderings, as shown in Fig.~\ref{fig:sample_diversity}.

\subsection{Adaptive Mask Inpainting}
\label{subsec:adaptivemaskinpainting}
The full pipeline of \textit{Adaptive Mask Inpainting} algorithm is described in Algorithm.~\ref{algo1:adaptivemaskinpaintingalgorithm}.
We use a publicly available inpainting diffusion model (RealisticVision~\cite{Rombach_2022_CVPR,realisticvision}) in our implementation, although we note that our adaptive mask algorithm can be applied to any inpainting diffusion model. We use classifier-free guidance scale of $11.0$, and apply DDIM~\cite{ddim} scheduler of $T=50$ timesteps with denoising strength set between $0.9\sim1.0$.

As the sequence of denoised image latents $\{x_t\}_{t=T}^0$ progresses (\emph{i.e.}, $t: T\xrightarrow[]{}0$), the quality of the predicted denoised image $\hat{x}_0$ improves over the progress of timestep, thus the low-level structure of the target prompt (\emph{i.e.}, human) becomes more apparent (as shown in Fig.~\ref{fig:adaptivemaskinpainting}). This allows us to ground the inpainting region for the next timestep ($m_{t-1}$) around that low-level structure by predicting the region using off-the-shelf segmentation model~\cite{pointrend}. Note that we use the initial inpainting mask ($m_{\text{default}}$) if the structure is not detected. We dilate the predicted human mask to tolerate the imperfectness of the generated structure during early steps, using $3 \times 3$ kernel $k$ with $n_t$ times repeat:
\begin{equation}
    k = 
    \begin{bmatrix}
        1 & 1 & 1 \\
        1 & 1 & 1 \\
        1 & 1 & 1
    \end{bmatrix}, \quad
    n_t = 
    \begin{cases}
        20 &\text{50 $\ge t$  $> 45$} \\
        10 &\text{45 $\ge t$  $> 40$} \\
        5 &\text{40 $\ge t$  $> 35$} \\
        4 &\text{35 $\ge t$  $> 30$} \\
        3 &\text{30 $\ge t$  $> 25$} \\
        2 &\text{25 $\ge t$  $> 20$} \\
        1 &\text{20 $\ge t$  $> 15$} \\
        0 &\text{15 $\ge t$  $> 0$} \\
    \end{cases}
\end{equation}

\noindent We also employ ``Provoke Schedule'' (refer to Algorithm.~\ref{algo1:adaptivemaskinpaintingalgorithm}) for faster generation speed. The provoke scheduler determines whether to skip the mask adaptation step ($t\in \text{ProvokeSchedule}$) or not ($t\not\in\text{ProvokeSchedule}$) during timestep $t$. Specifically, we use the following schedule:
\begin{equation}
    \text{ProvokeSchedule}= \{t~|~\text{(40 $\ge t$  $\ge$ 2 and $t$ is even) or $t$ = 45}\}
\end{equation}

\begin{algorithm}[t]
  \caption{Adaptive Mask Inpainting}
  \label{algo1:adaptivemaskinpaintingalgorithm}
  \begin{algorithmic}
    \STATE Latent Diffusion Model: $\epsilon_\theta$
    \STATE Latent VAE Decoder: $\mathbf{D}$
    \STATE Segmentation Model: $\mathbf{S}$
    \STATE DDIMSchedule: $\{\alpha_t\}_{t=1}^T$
    \STATE Dilation Schedule: $\{n_t\}_{t=1}^{T}$, Dilation Kernel: $k$
    \STATE Inputs: Prompt ($c$), Initial Mask ($m_{\text{default}}$), Image ($I_{\text{orig}}$) 
    \STATE Initialize Noise Latent: $x_T\sim \mathcal{N}(\mathbf{0},\mathbf{I})$
    \STATE Initialize Adaptive Mask: $m_T\xleftarrow[]{} m_{\text{default}}$
    
    \FOR{$t$=$T,...,1$}
        \STATE $\hat{x}_0 = {1 \over \sqrt{\bar\alpha_{t}}} (x_t - \sqrt{1-\bar\alpha_{t}}\epsilon_{\theta}(x_t;c,m_t,I_{\text{orig}},t))$, $\bar\alpha_{t}=\prod_{i=1}^{t}\alpha_i$
        \IF{$t\in \text{ProvokeSchedule}$}
            \STATE $s = \mathbf{S}(\mathbf{D}(\hat{x}_0))$
            \IF{$s\neq \varnothing$}
                \STATE $m_{t-1}=\text{Dilate}(s; n_t, k)$
            \ELSE
                \STATE $m_{t-1}=m_{\text{default}}$
            \ENDIF
        \ELSE
            \STATE $m_{t-1} = m_t$
        \ENDIF
        \STATE $x_{t-1} = \text{DDIMStep}(x_t, \hat{x}_0, t)$
    \ENDFOR
    \RETURN $\mathbf{D}(x_0)$
  \end{algorithmic}
\end{algorithm}

\subsection{Lifting 2D Affordance to 3D}
\label{subsec:liftingto3d}
\subsubsection{Number of Joints Used.} We use Hand4Whole~\cite{Moon_2022_CVPRW_Hand4Whole} to predict 3D humans from images ($\mathbf{F}_{\text{human}}$ in Eq.~\ref{eq:fhuman}). The model returns 1 global rotation (for pelvis joint) and 54 human joint rotations following the SMPL-X~\cite{SMPL-X:2019} format; which consists of $21$ body joints, ${15 + 15}$ hand joints, $1$ jaw joint, and $1+1$ eye joints. We add 21 OpenPose~\cite{openpose} joints (\textit{``nose'', ``right eye'', ``left eye'', ``right ear'', ``left ear'', ``left big toe'', ``left small toe'', ``left heel'', ``right big toe'', ``right small toe'', ``right heel'', ``left thumb'', ``left index'', ``left middle'', ``left ring'', ``left pinky'', ``right thumb'', ``right index'', ``right middle'', ``right ring'', ``right pinky''}) and exclude 11 original joints (\textit{``spine1'', ``spine2'', ``spine3'', ``left foot'', ``right foot'', ``left collar'', ``right collar'', ``head'', ``jaw'', ``left eye smplhf'', ``right eye smplhf''}), resulting in $1 + 54 + 21 - 11 = 65$ joints.

\subsubsection{Finding Inlier Set.}
We utilize a semi-consistent largest inlier set obtained from the generated 2D HOI image set $\{I_d\}_{d=1}^D$ to uplift the given generated 2D HOI image $I_{\text{ref}}$ to 3D. To find the inlier set, we first choose target image set $\mathcal{I}_{\text{target}}$ from $\{I_d\}_{d=1}^D$, consisting of images which is generated from different views and shows high consistency with reference image $I_{\text{ref}}$. Specifically, we first triangulate human joints for every pairs of $\{ I_{\text{ref}}, I_{\text{other}} \}$, where $I_{\text{other}} \in \{I_d\}_{d=1}^D$ is the image generated from different view with $I_{\text{ref}}$. We choose the best $N_{\text{triangulation}}$ images which the triangulated 3D human joints shows less re-projection error on reference image than threshold $\tau_{\text{triangulation}}$, constructing $\mathcal{I}_{\text{target}}$.

For every image $I_{\text{target}}$ in the target image set $\mathcal{I}_{\text{target}}$ and the corresponding 3D human joints obtained via triangulation of $I_{\text{ref}}$ and $I_{\text{target}}$, we find the number of inliers images which shows less re-projection error than $\tau_{\text{ransac}}$, denoted as $n_{\text{target}}$. We use the inlier set with maximum $n_{\text{target}}$ samples for the further depth optimization. In practice, we set $\tau_{\text{triangulation}} = 100$, $\tau_{\text{ransac}} = 100$, $N_{\text{triangulation}} = 400$, and use mean squared joint error on pixel space for all re-projection error.

\subsubsection{Initializing Depth.} We initialize 7 human candidates equispaced along the orthographic ray, where we place $4^{\text{th}}$ candidate (center candidate) to the position that minimizes the average distance between the pelvis joint and all object vertices. The distance between human candidates is proportional to the width of the human along the orthographic camera ray, where we set the multiplier as $0.3$. We initialize the depth using the human candidate with maximum IoU between the rendered human mask (regarding occlusion) and the predicted human segmentation mask. 

\subsubsection{Depth Optimization Settings.} For depth optimization, we set $\lambda_{\text{collision}}=400$, and use Adam~\cite{adam} optimizer with learning rate $1 \times 10^{-2}$ for $200$ iteration to optimize $\mathcal{L}$.

\subsubsection{Filtering.}
We filter out the 3D human samples if (1) the IoU between the human rendering and predicted human segmentation is below $0.3$ or over $0.8$, (2) number of inliers after RANSAC~\cite{fischler1981random} is below $\tau_{\text{inlier}}$ (which varies from $1\sim50$, based on the given 3D object), or (3) the intersection volume over human volume is higher than $0.01$.

\subsection{Learning Comprehensive Affordance}
\subsubsection{Canonicalization from $\mathbf{n}_j^h,\mathbf{p}^{o\xrightarrow[]{}h}$ to $\mathbf{n},\mathbf{p}$.} We address 2 types of 3D object (including 3D human): (1) the object is assumed rigid, meaning that current object can be obtained via applying rigid transformation $\mathbf{T}^{\text{original}\xrightarrow[]{}\text{current}}\in\text{SE}(3)$ on the original object; or (2) the object is non-rigid (\emph{e.g.}, 3D human), meaning that no original object exist, and also the rigid transformation. We provide the canonicalization procedure that addresses both cases. 

Given the human surface normal $\mathbf{n}_j^h$ and relative position $\mathbf{p}^{o\xrightarrow[]{}h}$, we rotate them the same amount when rotating the object surface normal $\mathbf{n}_i^o$ to face $\hat{\mathbf{n}}=[0,0,1]^T$
Specifically, we canonicalize the human normal $\mathbf{n}_j^h$ to $\mathbf{n}$ following:
\begin{equation}
    \label{eq:canonicalize_normal}
    \mathbf{n}=(\mathbf{n}_i^o\cdot \hat{\mathbf{n}})\mathbf{n}^h_j + (\mathbf{n}_j^h\cdot\mathbf{n}_i^o)\hat{\mathbf{n}} - (\mathbf{n}_j^h\cdot \hat{\mathbf{n}})\mathbf{n}_i^o + [{\mathbf{n}_j^h \cdot (\mathbf{n}_i^o\times \hat{\mathbf{n}})\over 1+\mathbf{n}_i^o\cdot\hat{\mathbf{n}}}] (\mathbf{n}_i^o\times \hat{\mathbf{n}})
\end{equation}
and similarly, we canonicalize the relative position $\mathbf{p}^{o\xrightarrow[]{}h}$ to $\mathbf{p}$ following:
\begin{equation}
    \label{eq:canonicalize_relative_length}
    \mathbf{p}=(\mathbf{n}_i^o\cdot \hat{\mathbf{n}})\mathbf{p}^{o\xrightarrow[]{}h} + (\mathbf{p}^{o\xrightarrow[]{}h}\cdot\mathbf{n}_i^o)\hat{\mathbf{n}} - (\mathbf{p}^{o\xrightarrow[]{}h}\cdot \hat{\mathbf{n}})\mathbf{n}_i^o + [{\mathbf{p}^{o\xrightarrow[]{}h} \cdot (\mathbf{n}_i^o\times \hat{\mathbf{n}})\over 1+\mathbf{n}_i^o\cdot\hat{\mathbf{n}}}] (\mathbf{n}_i^o\times \hat{\mathbf{n}})
\end{equation}
The canonicaliation procedure described in Eq.~\ref{eq:canonicalize_normal} and Eq.~\ref{eq:canonicalize_relative_length} preserves the length of the vector ($\|\mathbf{n}\|=\|\mathbf{n}^h_j\|$, $\|\mathbf{p}^{o\xrightarrow[]{}h}\|=\|\mathbf{p}\|$) and preserves the orientation with respect to the object normal ($\mathbf{n}_i^o\cdot\mathbf{n}_j^h=\hat{\mathbf{n}}\cdot\mathbf{n}$, $\mathbf{n}_i^o\cdot\mathbf{p}^{o\xrightarrow[]{}h}=\hat{\mathbf{n}}\cdot\mathbf{p}$). Also, the procedure above describes the movement of human normal $\mathbf{n}_j^h$ and relative position $\mathbf{p}^{o\xrightarrow[]{}h}$ following the object normal, when the object normal is taking the ``shortest path'' to $\hat{\mathbf{n}}$ along the sphere surface $\mathbb{S}^2$.
Note that for the case of rigid object, we transform the current object (and corresponding human) back to the original state using $(\mathbf{T}^{\text{original}\xrightarrow[]{}\text{current}})^{-1}$ before applying Eq.~\ref{eq:canonicalize_normal} and Eq.~\ref{eq:canonicalize_relative_length}. For non-rigid objects (\emph{e.g.}, human mesh), we directly apply Eq.~\ref{eq:canonicalize_normal} and Eq.~\ref{eq:canonicalize_relative_length}.

\subsubsection{Additional Details.} We sample $750\sim2048$ points from human mesh and object mesh using Poisson disk sampling~\cite{poissondisk2}. For human mesh, we find the closest vertex of SMPL-X~\cite{SMPL-X:2019} and save the vertex indices as the human mesh is not rigid and geometry may alter when the pose differs. 
We set the domain of $\mathbf{p}$ as $30\times30\times30$ voxelgrid with each voxel the size of $0.04$, and the domain of $\mathbf{n}$ as an equispaced spherical grid with 250 points, where the domain points are obtained via Fibonacci spirals~\cite{fibonaccispiral}. To save memory during quantitative evaluation (which only compares contact scores with previous approaches), we only accumulate $e^{-\|\mathbf{p}\|}$ instead of using full voxelgrid since $f_{\text{contact}}$ from Eq.~\ref{eq:contactaffordance} only requires relative distance $\|\mathbf{p}\|$ to compute. We fit the Gaussian kernel with $\sigma=0.2$ for the domain of $\mathbf{n}$, and $\sigma=0$ (quant) / $\sigma=0.1$ (qual) for the domain of $\mathbf{p}$. Note that the Gaussian kernel for $\mathbf{n}$ is computed using geodesic metrics, where we assume the radius of spherical grid as 1. Finally, we set $n_b=10^6$ when computing $f_{\text{orientation}}$ from Eq.~\ref{eq:implicitresponse}.

\section{Additional Details on Experiments}
\subsubsection{Preprocessing Intercap~\cite{InterCap}.}
Since Intercap~\cite{InterCap} did not provide texture for the 3D objects at the time of submission, we generated the texture for the objects using TEXTure~\cite{TexTure} where the stylization prompts are generated via ChatGPT~\cite{chatgpt} using the following query:
\begin{quote}
    \textit{Give a simple appearance description of an object of given categories as a form of ``a }$\{\text{\textbf{category}}\}$, $\{\text{\textbf{appearance description}}\}$''.
\end{quote}

\subsubsection{Method for Aggregating Contact Maps.} When aggregating $N$ samples to compute $\mathcal{P}_{ij}$'s for all pairs of $i$-th object point and $j$-th human point, we also count the number of times when $\|\mathbf{p}\|<d_{\text{thres}}$, which we denote as $N_{\text{sig}}^{ij}$. To create a holistic contact map, we aggregate the contact values derived from $\mathcal{P}_{ij}$ only if $N_{\text{sig}}^{ij}/N>\tau_{\text{sig}}$, assuming such $ij$ pairs show significant contact. When aggregating the contact values, we simply choose the maximum value between $ij$ pairs. We set $d_{\text{thres}}=0.1$, $\tau_{\text{sig}}=0.05$ in our implementation.   

\begin{figure}[t]\centering
\includegraphics[width=0.96\linewidth, trim={0 0 0 0},clip]{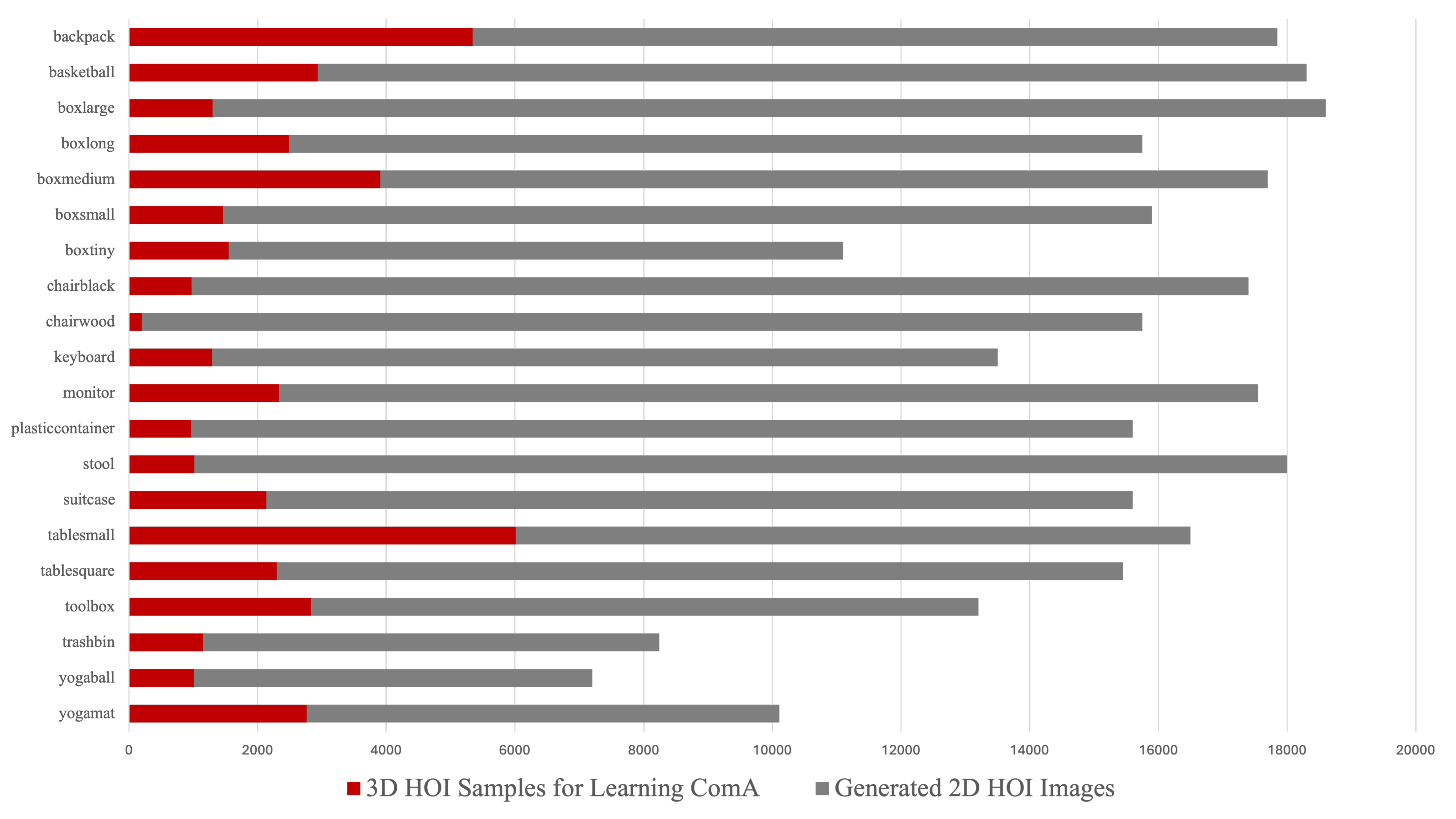}
\vspace{-5pt}
\captionof{figure}{\textbf{Statistics on Generated Samples.} We report the number of generated 2D HOI images (gray) and 3D HOI samples (red) for BEHAVE objects, which we use in both qualitative and quantitative evaluation.
}
\label{fig:statistics}
\vspace{-20pt}
\end{figure}

\begin{figure}[p!]\centering
\includegraphics[width=\linewidth, trim={0 0 0 0},clip]{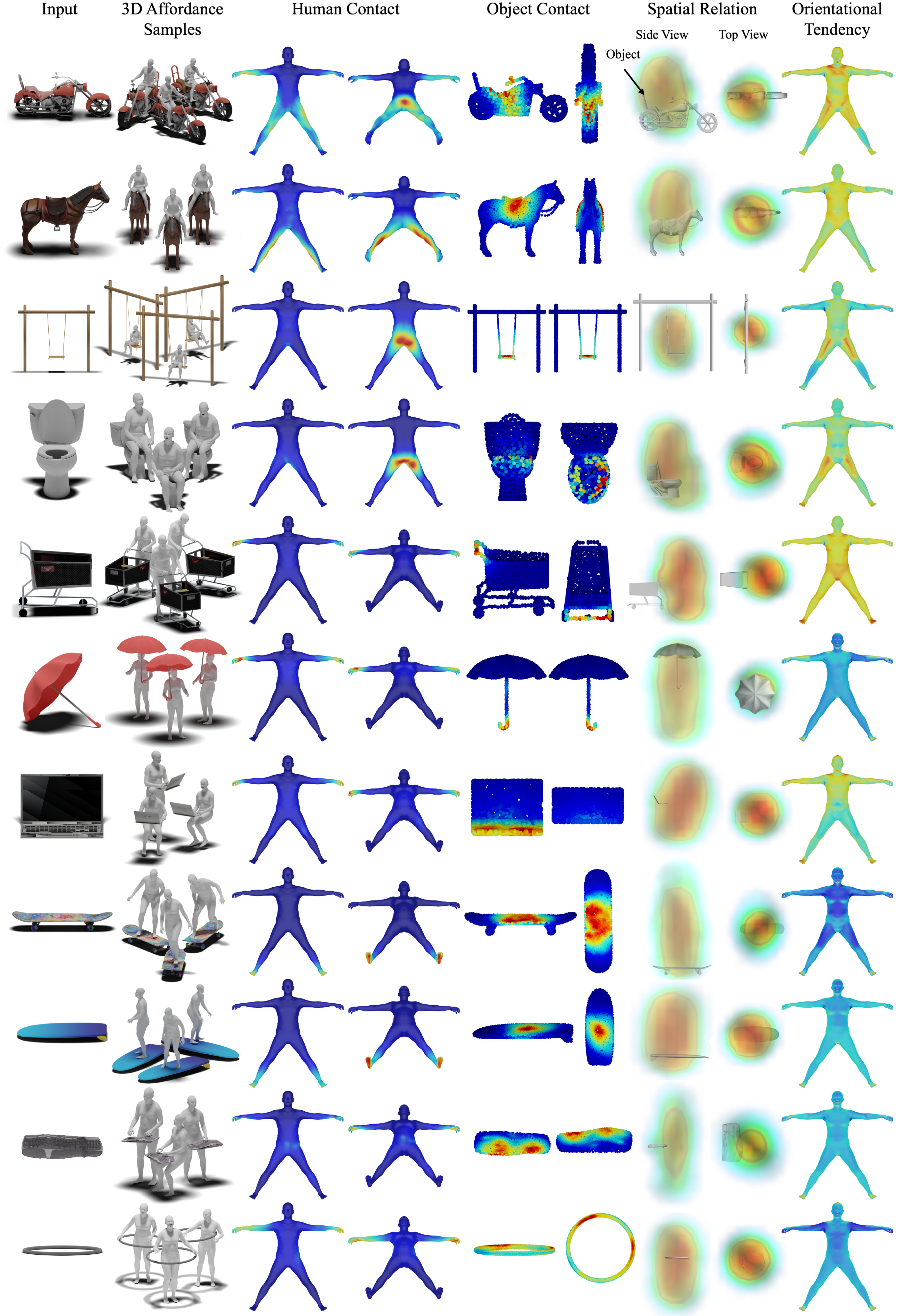}
\vspace{-5pt}
\captionof{figure}{\textbf{Additional Qualitative Results.}
Our method can be applied to various 3D objects obtained from diverse sources.
}
\label{fig:additional_qual}
\vspace{-20pt}
\end{figure}

\subsubsection{Statistics of 2D HOI Images and 3D HOI Samples} We report statistics on the generated 2D HOI images and 3D HOI samples. Fig.~\ref{fig:statistics} presents statistics on the BEHAVE~\cite{BEHAVE} dataset objects, which were used for both qualitative and quantitative evaluation. We generate images with varying mask regions, text prompts, and random seeds, resulting in a minimum of 7200, a maximum of 18600, and an average of 14965.15 images. After filtering out malicious data, we finally obtain a minimum of 198, a maximum of 6011, and an average of 2198.5 3D HOI samples for learning ComA. The overall acceptance ratio is 14.69\%, meaning that we need approximately 7 images to obtain a single 3D HOI sample.

\subsubsection{Additional Qualitative Results.} We report additional qualitative results in Fig.~\ref{fig:additional_qual}. As we utilize the affordance knowledge inherent in pre-trained 2D diffusion models, we are able to learn ComA for uncommon categories (\emph{e.g.}, horse, swing, toilet, cart) which are not typically addressed in traditional 3D HOI datasets~\cite{BEHAVE, InterCap}.

\subsubsection{Details on Application.} We use SMPL-X~\cite{SMPL-X:2019} model to optimize global orientation, translation, body pose, and hand pose. For the body pose, we optimize pose embedding following VPoser~\cite{SMPL-X:2019} to leverage pose prior loss $\mathcal{L}_{\text{pprior}}$ and angle prior loss $\mathcal{L}_{\text{aprior}}$ which helps generating plausible pose. We define orientation loss $\mathcal{L}_{\text{orientation}}$ as average normalized cosine similarity between maximum probability direction (while object is fixed) obtained from ComA and human vertex normal for all human vertices. We also define contact loss $\mathcal{L}_{\text{contact}}$ as a chamfer distance between each contact points of human and object obtained from ComA. The total loss is defined as below:

\begin{equation}
    \mathcal{L}_{\text{opt}} = \lambda_1 \mathcal{L}_{\text{pprior}} + \lambda_2 \mathcal{L}_{\text{aprior}} + \lambda_3 \mathcal{L}_{\text{orientation}} + \lambda_4 \mathcal{L}_{\text{contact}}
\end{equation}

\noindent In practice, we use $\lambda_1 = 1\times 10^{-6}$, $\lambda_2 = 3.17\times 10^4$, $\lambda_3 = 1\times 10^{12}$, and $\lambda_4 = 2.6\times 10^{11}$, and optimize $\mathcal{L}_{\text{opt}}$ for 2000 steps using Adam~\cite{adam}.

\section{Limitations \& Future Works}

\noindent \textbf{Spatial Bias in Inpainting Diffusion Models.} Our method utilizes inpainting diffusion models~\cite{Rombach_2022_CVPR, realisticvision} to insert humans into object images; however, the diffusion model may possess spatial biases, which may alter the inpainting results as the properties of inpainting mask (\emph{e.g.}, center location, aspect ratio, resolution) differs. For example, diffusion model may not be able to generate humans if the objects that usually interact with hands (\emph{e.g.}, sports ball) are rendered on the bottom side of the image. Future research can further improve by mitigating the spatial bias in diffusion models, or the mask selection procedure to reduce the number of unnecessary generations.

\noindent \textbf{Incorrect HOI Prompt Generation.} During the prompt generation step, there is a chance for the vision-language model~\cite{gpt4v} to misidentify the object in the image, resulting in incorrect HOI prompts that describe implausible situations for the given object.

\noindent \textbf{Limits and Potentials of Adaptive Mask Inpainting.} We propose an \textit{Adaptive Mask Inpainting} algorithm to preserve the original object during inpainting. However, the method depends on a segmentation model to adapt the inpainting mask, and the errors during segmentation may affect the inpainting results. For example, if the segmentation model predicts part of the object as human due to various reasons (\emph{e.g.}, the texture of the object is similar to the texture of the generated human), the algorithm may not work well as the following inpainting mask also occludes the object. One potential approach for improvement is to use better segmentation models, such as Grounded SAM~\cite{liu2023groundingdino,Kirillov_2023_ICCV_SegmentAnything}. 

While we use adaptive mask inpainting only for the human insertion task, the algorithm can be applied to any categories the segmentation model allows, opening possibilities such as \textit{open-vocabulary object insertion into scene image}.

\noindent \textbf{Bias due to Filtering.} Employing heavy filtering at the end of the pipeline may result in bias. For example, filtering out humans with high collision may cause the remaining samples to ``slide out'' the object, especially if the object is complex and is highly likely to collide even when given the plausible posture (\emph{e.g.}, motorcycle). One alternative is applying soft filtering (\emph{i.e.}, applying confidence weights instead of removing images with hard thresholds).

\noindent \textbf{Large Memory Consumption.} ComA returns distributions for each pair of human and object points, which leads to large memory consumption when the resolution of human and object mesh is high; forcing us to downsample the human and object mesh. The limited resolution may cause the representation to lack details, especially when modeling interactions with dexterous objects. It is worth exploring the use of implicit 3D representation for human and object surfaces (\emph{e.g.}, SDF~\cite{oleynikova2016signeddistancefields}, DMTet~\cite{shen2021dmtet}), as such representations model the continuous surface as function.

\begin{figure}[t]
\centering
\includegraphics[width=0.85\columnwidth]{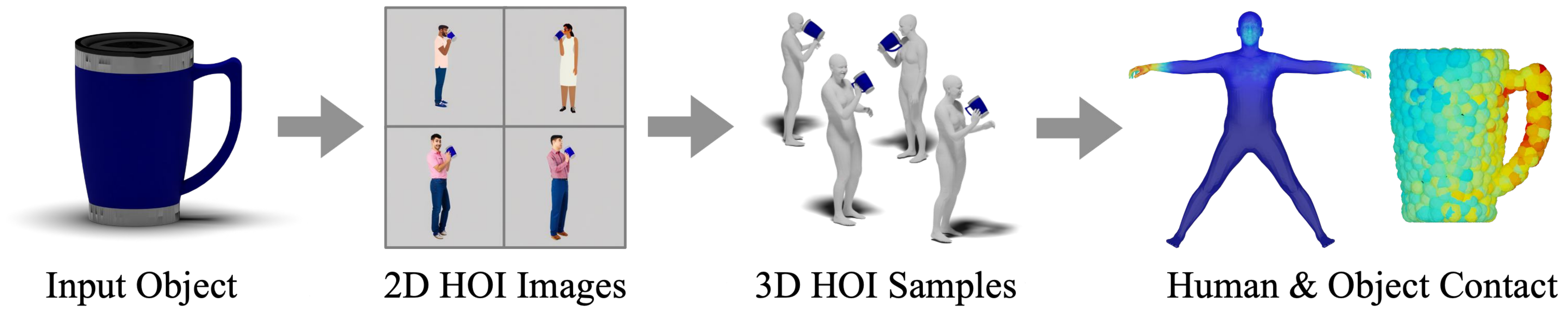}
\vspace{-5pt}
\caption{\textbf{Results of ComA for Small Object.} While our pipeline captures plausible affordance even for small objects (such as cup), the results show low granularity compared to big objects.}
\label{fig:small_object}
\vspace{-15pt}
\end{figure}

\noindent \textbf{Low Granularity for Small Objects.} Although our pipeline captures affordance even for small objects (\emph{e.g.}, cup interacting with hand and mouth, as shown in Fig.~\ref{fig:small_object}), the lack of granularity compared to big objects is an existing challenge to solve.

\noindent \textbf{Modelling Hand-Object Interactions.} 
Diffusion models often struggle to produce high-quality images of hands. Future research could benefit from using diffusion models trained specifically on hand images to improve hand generation and employing close-view cameras focused on hands for better modeling.

\noindent \textbf{Possible Improvements in ComA.}
We introduce ComA as a new representation for affordances and use it to deduce contact and non-contact information. Although our method for deriving contact is well-founded, there are opportunities for enhancement. Specifically, incorporating pressure modeling could extend ComA's applicability to deformable objects. Additionally, the concept of orientational affordance needs refinement. The current approach effectively measures orientation preference using a negated entropy term, but it fails to identify the underlying reasons for this preference. For instance, in the case of chair, while human feet often orient towards the ground, attributing this tendency to the object itself overlooks the influence of gravity. A valuable future research direction would involve distinguishing and analyzing the reasons behind orientational tendencies.

\noindent \textbf{Expanding to Non-Watertight Mesh.} ComA can be easily extracted from non-rigid mesh (\emph{e.g.}, human mesh), as long as the mesh provides a closed surface and surface normal can be defined. One possible future direction is to improve ComA to be easily extractable from any 3D surfaces (mesh or other surface representations), including non-watertight mesh, and sharp meshes where defining surface normal is non-trivial.

\noindent \textbf{Evaluation Metrics.} 
There's potential to explore more complex metrics to accurately assess the effectiveness of our method, particularly when evaluating the volumetric quality of 3D humans and objects to support the use of 2D-to-3D conversion methods.

\begin{figure}[t]
\centering
\includegraphics[width=0.75\columnwidth]{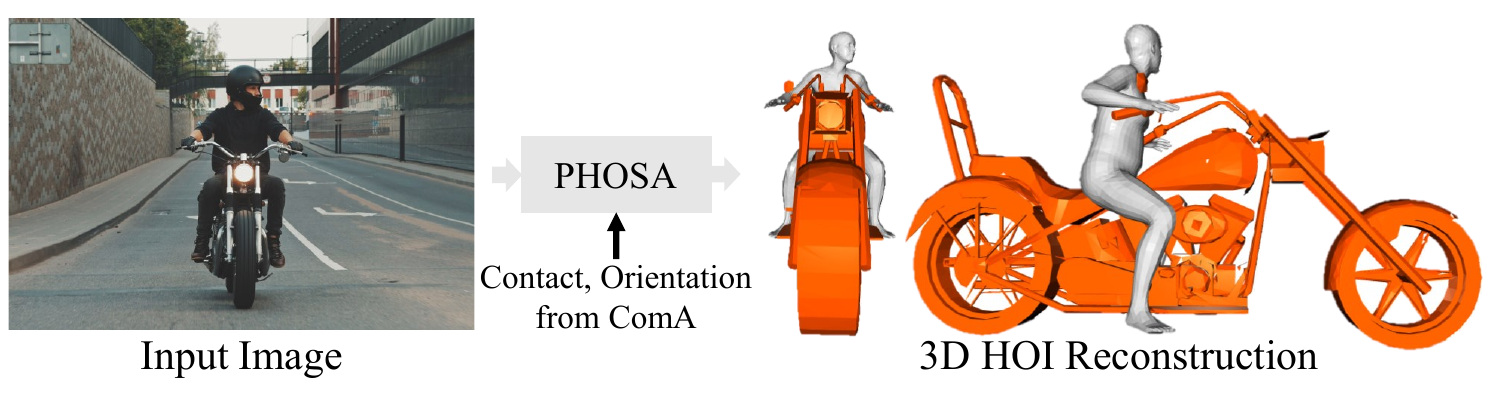}
\vspace{-5pt}
\caption{\textbf{Single Image HOI Reconstruction for Any Object using ComA.} We can directly replace the contact heuristics in PHOSA~\cite{phosa} and apply additional orientation loss from ComA, which allows PHOSA~\cite{phosa} to scale to unseen objects.}
\label{fig:phosapluscoma}
\vspace{-15pt}
\end{figure}

\noindent \textbf{Potential Applications.}
Our new method and ComA provides multitudes of possible applications, as demonstrated in Sec.~{4.5}. We list potential downstream applications: (1) Large-scale 3D affordance dataset generation; (2) Single image HOI reconstruction for any 3D object (as shown in Fig.~\ref{fig:phosapluscoma}); (3) Object recognition from 3D human posture (similar to Object Pop-up~\cite{objectpopup}); (4) Action recognition from human-object interaction sequence; (5) Application for robotics, especially for humanoids.

\end{document}